\documentclass[nohyperref]{article}

\usepackage{amsmath,amsfonts,bm}

\def\eqref#1{equation~\ref{#1}}
\def\Eqref#1{Equation~\ref{#1}}

\def\1{\bm{1}}

\def\vh{{\bm{h}}}
\def\vi{{\bm{i}}}

\def\vv{{\bm{v}}}

\def\vx{{\bm{x}}}

\def\vz{{\bm{z}}}

\def\mA{{\bm{A}}}

\def\mW{{\bm{W}}}
\def\mX{{\bm{X}}}

\DeclareMathAlphabet{\mathsfit}{\encodingdefault}{\sfdefault}{m}{sl}
\SetMathAlphabet{\mathsfit}{bold}{\encodingdefault}{\sfdefault}{bx}{n}

\def\gG{{\mathcal{G}}}

\def\sN{{\mathbb{N}}}

\def\sP{{\mathbb{P}}}

\def\sR{{\mathbb{R}}}

\def\sZ{{\mathbb{Z}}}

\usepackage{microtype}
\usepackage{graphicx}
\usepackage{subfigure}
\usepackage{booktabs} 
\usepackage{arydshln}
\usepackage{makecell}
\usepackage{float}

\usepackage{hyperref}

\usepackage{tabularx}

\usepackage[accepted]{icml2023}

\usepackage{amsmath}
\usepackage{amssymb}
\usepackage{mathtools}
\usepackage{amsthm}
\usepackage{amsfonts}
\usepackage{booktabs}       
\usepackage{multicol}
\usepackage{multirow}

\usepackage[capitalize,noabbrev]{cleveref}

\theoremstyle{plain}
\newtheorem{theorem}{Theorem}[section]

\newtheorem{lemma}[theorem]{Lemma}

\theoremstyle{definition}

\newtheorem{assumption}[theorem]{Assumption}
\theoremstyle{remark}

\crefformat{equation}{(#2#1#3)}
\crefname{assumption}{Assumption}{Assumptions}

\usepackage[textsize=tiny]{todonotes}

\icmltitlerunning{On the Initialization of Graph Neural Networks}

\begin{document}

\twocolumn[
\icmltitle{On the Initialization of Graph Neural Networks}

\icmlsetsymbol{equal}{*}

\begin{icmlauthorlist}
\icmlauthor{Jiahang Li}{polyu,equal}
\icmlauthor{Yakun Song}{sjtu,equal}
\icmlauthor{Xiang Song}{awsai}
\icmlauthor{David Paul Wipf}{asail}
\end{icmlauthorlist}

\icmlaffiliation{polyu}{The Hong Kong Polytechnic University}
\icmlaffiliation{sjtu}{Shanghai Jiao Tong University}
\icmlaffiliation{awsai}{Amazon AI}
\icmlaffiliation{asail}{Amazon Shanghai AI Lab}

\icmlcorrespondingauthor{Jiahang Li}{jiahangspongebob.li@connect.polyu.hk}
\icmlcorrespondingauthor{Xiang Song}{xiangsx@amazon.com}

\icmlkeywords{Graph Neural Networks, Initialization}

\vskip 0.3in
]

\printAffiliationsAndNotice{*Work was done during an internship in Amazon Shanghai AI Lab}

\newcommand{\david}[1]{{ \color{red}{[David: #1]} }}

\definecolor{amber}{rgb}{1.0, 0.49, 0.0}
\newcommand{\jiahang}[1]{{ \color{amber}{[Jiahang: #1]} }}
\newcommand{\method}{Virgo}

\begin{abstract}
Graph Neural Networks (GNNs) have displayed considerable promise in graph representation learning across various applications.
The core learning process requires the initialization of model weight matrices within each GNN layer, which is typically accomplished via classic initialization methods such as \textit{Xavier} initialization. 
However, these methods were originally motivated to stabilize the variance of hidden embeddings and gradients across layers of Feedforward Neural Networks (FNNs) and Convolutional Neural Networks (CNNs) to avoid vanishing gradients and maintain steady information flow. 
In contrast, within the GNN context classical initializations disregard the impact of the input graph structure and message passing on variance. 
In this paper, we analyze the variance of forward and backward propagation across GNN layers and show that the variance instability of GNN initializations comes from the combined effect of the activation function, hidden dimension, graph structure and message passing. To better account for these influence factors, we propose a new initialization method for \textbf{V}ariance \textbf{I}nstability \textbf{R}eduction within \textbf{G}NN \textbf{O}ptimization (Virgo), which naturally tends to equate forward and backward variances across successive layers.
We conduct comprehensive experiments on 15 datasets to show that \method\ can lead to superior model performance and more stable variance at initialization on node classification, link prediction and graph classification tasks. Codes are in \href{https://github.com/LspongebobJH/virgo_icml2023}{Virgo}. 
\end{abstract}
\section{Introduction}
Graph Neural Networks (GNNs)~\cite{graphsage,gat,gcn,gin,monet} are a class of deep learning models specifically designed to process and analyze graph-structured data, allowing for the integration of both node and edge information for tasks such as node classification, link prediction, and graph classification. GNNs have recently shown great success in graph representation learning in service of various downstream applications including social networks~\cite{ying2018graph,rossi2020temporal}, recommendation~\cite{fan2019graph,yu2021self}, fraud detection~\cite{wang2019fdgars,liu2020alleviating}, and life sciences~\cite{strokach2020fast,jing2021fast,nguyen2020predicting}.

To initiate model training, learnable GNN weight matrices need to be initialized in one way or another.  In the past, more traditional deep learning models (e.g., CNNs, MLPs) have typically adopted initialization schemes designed to improve training outcomes by stabilizing the variance of forward and backward passes \cite{lecun,xavier,kaiming}, the motivation being that unstable variances could otherwise lead to undesirable phenomena such as vanishing gradients~\cite{lecun} or poor information flow~\cite{xavier}.   The GNN community has largely borrowed these same schemes, particularly the \textit{Xavier}~\cite{xavier} and \textit{Lecun}~\cite{lecun} initialization paradigms. For example, the Deep Graph Library (DGL)~\cite{dgl} uses \textit{Xavier} to initialize the layers of GNN architectures such as GCN~\cite{gcn}, GraphSAGE~\cite{graphsage}, GAT/GATv2~\cite{gat,gatv2} and SGC~\cite{sgc} models. Similarly, the PyTorch Geometric (PyG) package~\cite{pyg} also uses \textit{Xavier} to initialize models like GCN, GAT/GATv2, and RGCN~\cite{rgcn}. Meanwhile, RGCN and ChebNet~\cite{chebnet} layers within DGL and GraphSAGE layers within PyG adopt \textit{Lecun} initialization.


And yet despite this widespread adoption within GNN training, it remains unclear the degree to which the original justifications for existing initializations actually still apply when we venture beyond the non-GNN architectures for which they were first designed.  Indeed, prior analysis has largely relied on the assumption of i.i.d.~training instances devoid of graph structure (and all neurons within each layer having the same variance), which greatly simplifies the selection of an appropriate distribution for drawing initial matrices.  The latter is usually a zero-mean uniform or Gaussian distribution with variance chosen to reduce the influence of the model hidden dimension~\cite{lecun,xavier,kaiming} or activation function~\cite{kaiming} on the variance.  However, GNN message passing layers and graph structure can impact initial variances in more nuanced ways, for example, through dependencies introduced by varying node-wise receptive-field sizes.  Hence prior assumptions may no longer apply and it behooves us to consider GNN-specific alternatives.


For this purpose, we first present derivations for forward and backward variances within a certain class of message passing GNNs. Specifically, for any given layer, we decompose the variance of each node into the sum of variances over message propagation paths, and then further decompose the variance of each message propagation path into the sum of variances over weight propagation paths. As a result of these cascaded decompositions, we obtain expressions for the variance of each node in terms of the variance of weight matrices. These expressions disclose the combined impact of hidden dimension, activation function, graph structure, and message aggregation mechanism of the variance of hidden embeddings and gradients.

Based on these insights, we next propose a simple but effective initialization method called \textit{Virgo} for \textbf{V}ariance \textbf{I}nstability Reduction within \textbf{G}NN \textbf{O}ptimization to mitigate the influence of these factors. Defining the overall variance within each layer as the mean over the variances of all nodes, \method\ minimizes the difference of overall variance between successive layers, and thereby derives the variance of distributions, such as zero-mean Gaussian or uniform, from which we can sample initial weight matrices for GNNs. Finally, we conduct comprehensive experiments on 15 datasets across three popular graph tasks, namely, node classification, link prediction, and graph classification. We compare \method\ with existing initialization methods including \textit{Lecun}, \textit{Xavier} and \textit{Kaiming}. Overall, we make following contributions:
\begin{itemize}
    \item We derive and analyze expressions for the variance of GNN embeddings (forward pass) and gradients (backward pass), showcasing how these quantities are affected by the joint influences of hidden dimension, activation function, graph structure, and GNN message passing mechanisms.

    \item  We propose a new initialization method named Virgo for GNN weight matrices based on our analysis, which minimizes the difference of the overall variance between successive layers. 

    
    \item We evaluate GNNs with different initializations on node classification, link prediction and graph classification tasks. \method\ helps improve prediction accuracy by up to 7\% and well stabilizes variances at the initialization.
\end{itemize}

\section{Preliminaries}
This section introduces basic GNN concepts and initialization methods. And for convenience, we summarize our adopted notational conventions in~\cref{tab:nota}.

\begin{table*}[hbpt]
    \centering
    \small
    \caption{Notation table. }
    
    \vskip 0.1in
    \begin{tabular}{cl}
    \toprule
      $\mW^l$ & Weight matrix of the $l$-th GNN layer, $\mW^{l} \in \sR^{m_1^{(l)} \times m_2^{(l)}}$. $\mW^{l}_{ij}$ is denoted by $w_{i,j}^l$ \\
      \midrule
      $\hat \vh_i^l$, $\vh_i^l$ & \makecell[l]{Embedding of node $i$ at the $l$-th layer before and after the activation. $\vh_i^0$ is input feature of node $i$} \\
      \midrule
      $i$, $j$ & Node index\\
      $t$ & Neuron index\\
      $l$, $L$ & Layer index and number of layers of a GNN\\
      $h_{i,t}^l$, $w_{t_1,t_2}^l$ & The $t$-th element of embedding of node $i$ at $l$-th layer, and ($t_1$, $t_2$)-th element of $\mW^l$\\
      $d_{ij}$ & Re-normalization coefficient between node $i$ and $j$ following the definition of GCN~\cref{eq:gcn_for_ori}\\
      
      \midrule
      $\sN$, $\sN(i)$ & Set of nodes in $\gG$ and one-hop neighbors of node $i$ \\
      $[\vh_i^l]_p$ & \makecell[l]{Message propagation path indexed by $p$. This path has length $l$ and takes $i$ as the destination \\node. The set of all such $p$ is denoted as $\sP_{i,l}$}\\
      $[h_i^l]_{p,t,\phi}$ & \makecell[l]{Weight propagation path indexed by $\phi$. This path goes along the message propagation path $p$, is \\connected to the $t$-th neuron of node $i$ and has length $l$. The set of all such $\phi$ is denoted as $\Phi_{i,p,t}$}\\
      
      \midrule
      $\delta(h)$, $\delta(\vh)$ & \makecell[l]{$\delta(h)$ is an indicator function of $h$, which is equal to 1 when $h$ is greater than 0, \\and 0 when $h$ is smaller than or equal to 0. $\delta(\vh)$ is a vector, of which elements are $\delta(h)$}\\
      \midrule
      $\odot$, $\prod_{\odot}$ & Element-wise product and cumulative element-wise product\\
      \bottomrule
     \end{tabular}
     \label{tab:nota}
    \end{table*}
    
\subsection{Graph Neural Networks}
\label{subsec:gcns}

Given a graph $\gG = (\mX, \mA)$, where $\mX$ is the feature matrix of nodes, $\mA$ is the graph adjacency matrix, the forward propagation over $\gG$ of the $l$-th GNN layer can be defined as:
\begin{equation}
    \vh_i^l = \sigma\left(\sum_{j \in \sN(i)}d_{ij}\vh_j^{l-1} \mW^{l-1} \right).
    \label{eq:gcn_for_ori}
\end{equation}
In this expression, $\sN(i)$ is a set including 1-hop neighbors of node $i$, $\sigma$ is an activation function, which we assume to be ReLU in this paper, $\mW^l \in \sR^{m_1^{(l)} \times m_2^{(l)}}$ is a weight matrix, and $\vh_i^l$ denotes the hidden embedding of the $i$-th node. We also set the scaling constant $d_{ij}$ to $1/\sqrt{(d_i+1)(d_j+1)}$ following the GCN model~\cite{gcn}, where $d_i$ and $d_j$ are the degrees of node $i$ and $j$ respectively. We denote the $t$-th element of the hidden embedding $\vh_i^l$ by $h_{i,t}^l$. $\bar{\vh}_i^l$ and $var(h_i^l) = \mathrm{Var_t}(h_{i,t}^l)$
denote the mean and variance over neurons within $\vh_i^l$ respectively. We use $var(h^l)$ to denote the mean over $var(h_i^l)$ of all nodes $i$. $var(h^l)$ is also termed as the \textbf{forward variance} of the $l_{th}$ layer. Analogously, the mean over $var(\frac{\partial Loss}{\partial h_i^l})$ of all nodes $i$, denoted by $var(\frac{\partial Loss}{\partial h^l})$, is termed as the \textbf{backward variance} of the $l_{th}$ layer, where $Loss$ refers to a standard cross entropy objective we assume throughout the paper. Hereinafter we use \textbf{variance} to denote these two types of variance without ambiguity. 

At initialization, we assume that neurons within $\vh_i^l$ are independent and identically distributed (i.i.d). The same assumption also applies to elements of $\mW^l$. Analogously, we assume that for all neurons $t$, $t'$, two certain nodes $i$, $j$ and a layer $l$, $\frac{\partial h_{j,t'}^L}{\partial h_{i,t}^l}$ are i.i.d. We also assume that elements of $\mW^l$ are independent to elements of $\mW^{l'}$, where $l$ is not equal to $l'$. The input features $\vh_i^0$ are random variables and the following derivations are conditioned on the input features of the given graph $\gG$.

\subsection{Classic Initializations Used by GNNs}

We denote the $(i,j)$-th of the weight matrix $\mW^l$ by $w_{i,j}^l$. In the following, we remove subscripts $(i,j)$ without causing ambiguity. Before the model training, elements of the weight matrix $\mW^l$ are sampled from a probability distribution $P(w^l)$, such as a uniform or Gaussian distribution. The mean of the distribution is typically set to zero, and thus the variance, denoted by $var(w^l)$, determines the form of the distribution. Classic initialization methods, such as \textit{Xavier} and \textit{Lecun}, tend to stabilize variance across layers by setting an appropriate $var(w^l)$ for all layers. Stabilizing variance means equating $var(h^l)$ for $\forall l \in {0 \cdots L-1}$, and equating $var(\frac{\partial Loss}{\partial h^l})$ for $\forall l \in {0 \cdots L-1}$. In the context of CNNs, $h^l$ denotes an element of a flattened feature map of the $l_{th}$ layer. For example, to stabilize forward variance, \textit{Lecun}, \textit{Xavier} and \textit{Kaiming} initialization set $var(w^l)$ to $\frac{1}{3m_1^{(l)}}$, $\frac{1}{m_1^{(l)}}$ and $\frac{2}{m_1^{(l)}}$ respectively. Meanwhile, \textit{Xavier} and \textit{Kaiming} initialization set $var(w^l)$ to $\frac{1}{m_2^{(l)}}$ and $\frac{2}{m_2^{(l)}}$ respectively to stabilize backward variance. Note that \textit{Xavier} sets the final $var(w^l)$ to the harmonic average of $\frac{1}{m_1^{(l)}}$ and $\frac{1}{m_2^{(l)}}$, obtaining a trade-off between forward and backward variance stabilization. 
Classic initializations might lead to suboptimal performance for GNNs though they have been widely used by modern GNN architectures, since they disregard the impact of input graph structure and message mechanisms of GNN on variance. Furthermore, they implicitly assume that forward outputs and backward gradients of all neurons within each layer have the same variance. We next derive new variance expressions that directly take graph structure into account.
\section{Forward and Backward Variance}
\label{sec:FBV}
In this section, we show an overview of derivations that finally provide analytic expressions of forward variance of GNNs defined by~\cref{eq:gcn_for_ori}. The backward variance follows similar derivations. Then we present variance expressions in terms of $var(w^l)$ based on given derivations. Proofs of theorems and empirical verification of assumptions are presented in the \cref{appendix}.

Firstly, to simplify subsequent derivations, we adopt a modification, proposed by~\cite{jknet}, to the activation function in the original GCN formulation~\cref{eq:gcn_for_ori}. Given an arbitrary vector $\vx$, we introduce an indicator function $\delta(\mathbf{x})$ that maps $\mathbf{x}$ into a binary vector, where each element is 1 if the corresponding element in $\mathbf{x}$ is greater than 0, and 0 otherwise. $\vz = \sigma(\mathbf{x})$ is thereby rewritten by $\vz = \delta(\vx) \odot \vx$. In this paper, we originally intend to investigate the influence of $\sigma(\mathbf{x})$ on $\vz$. The nesting of the activation function and the input vector makes such investigation intractable. After the proposed modification decouples the nesting, we are able to investigate the influence of $\delta(\vx)$ and $\vx$ on $\vz$ separately, where $\delta(\vx)$ is a simple 0-1 vector. We therefore convert~\cref{eq:gcn_for_ori} into the following equations:
\begin{equation}
\begin{aligned}
    \hat \vh_i^{l-1} &= \sum_{j \in \sN(i)}d_{ij}\vh_j^{l-1} \mW^{l-1}\\
    \vh_i^l &= \delta(\hat \vh_i^{l-1}) \odot \hat \vh_i^{l-1}.
\label{eq:gcn_for}
\end{aligned}
\end{equation}

\subsection{The First Variance Decomposition}
Based on~\cref{eq:gcn_for}, we now decompose the forward variance into the sum over variance of message propagation paths. As proposed by~\cite{jknet,influence}, we can decompose the acyclic GNN computation graph into a set of message propagation paths from the input to the $l$-th layer such that the embedding $\vh_i^l$ can be viewed as a summation over message propagation paths of length $l$. The full details of how we conduct the decomposition are presented in~\cref{misc}.

We denote an ordered sequence describing nodes along a message propagation path from node $i$ to node $j_{l-1}$ by $p = \{i, j_1, j_2, \cdots j_l\}$. The product set of neighbors $\{\sN(i)\times \sN(j_1) \cdots \sN(j_{l-2}) \times \sN(j_{l-1})\}$ is denoted by $\sN(i, j_{l-1})$. The decomposition results in:
\begin{small}
\begin{equation}
\begin{aligned}
    \vh_i^{l}=\sum_{p \in \atop \sN(i, j_{l-1})} &[(\prod_{k_1=0 \atop \odot}^{l-1}\delta(\hat \vh_{j_{k_1}}^{l-k_1-1}))(\prod_{k_2=0}^{l-1}d_{j_{k_2}j_{k_2+1}})\\
    &(\vh_{j_l}^0\prod_{k_3=0}^{l-1}\mW^{k_3})].
    \label{eq:mpath_ori}
\end{aligned}
\end{equation}
\end{small}

We denote a message propagation path of length $l$ as $[\vh_i^l]_{p}$, which is an element of the summation in~\cref{eq:mpath_ori}. $[\vh_i^l]_{p}$ is equal to:
\begin{small}
\begin{equation}
[\vh_i^l]_{p}=(\prod_{k_1=0 \atop \odot}^{l-1}\delta(\hat \vh_{j_{k_1}}^{l-k_1-1}))(\prod_{k_2=0}^{l-1}d_{j_{k_2}j_{k_2+1}})(\vh_{j_l}^0\prod_{k_3=0}^{l-1}\mW^{k_3}).
\label{eq:mpath}
\end{equation}
\end{small}

$[\vh_i^l]_{p}$ propagates input message from a $l$-hop neighbor $j_l$ to the target node $i$ along the path $p$. We denote the set including all such paths $p$ as $\sP_{i,l}$. Let $[h_i^l]_{p,t}$ be an element of the $[\vh_i^l]_{p}$, then $var(h_i^l)$ is equal to $var(\sum_{p \in \sP_{i,l}} [h_i^l]_{p,t})$, that is, the variance of node $i$ is equal to the variance of the sum over all message propagation paths into node $i$. 
We know that $var(\sum_{p \in \sP_{i,l}} [h_i^l]_{p,t})$ is equal to $\sum_{p \in \sP_{i,l}} var([h_i^l]_{p,t}) + 2\sum_{p_1 \ne p_2} cov([h_i^l]_{p_1,t}, [h_i^l]_{p_2,t})$.  We convert the covariance terms into the combination of variance terms $var([h_i^l]_{p,t})$ over different paths $p$, so that the variance of each node can be deduced from the variance of its message propagation paths. 
We observe that $[h_i^l]_{p_1,t}$ and $[h_i^l]_{p_2,t}$ have the multiplication of the same weight matrices, and $\delta$ terms only control whether or not a message propagation path is activated. Motivated by this observation, we make the following assumption to simplify the derivation and ease the conversion: 

\begin{assumption}
\label{assu:1}
Let $p_1$ and $p_2$ be two different elements of $\sP_{i,l}$. Before model training, the Pearson correlation between $[h_i^l]_{p_1,t}$ and $[h_i^l]_{p_2,t}$ is approximately 1.
\end{assumption}
In~\cref{appendix}, we take the GNN following~\cref{eq:gcn_for} and empirically investigate Pearson correlation of a set of message propagation paths on four datasets. All correlation results are greater than 0.83, which supports the feasibility of~\cref{assu:1} for GNNs.

Based on the \cref{assu:1}, the covariance term can be approximated by $\sqrt{var([h_i^l]_{p_1,t})var([h_i^l]_{p_2,t})}$, $var(\sum_{p \in \sP_{i,l}} [h_i^l]_{p,t})$ can thus be represented in terms of the combination of $var([h_i^l]_{p,t})$ over different paths $p$.

\subsection{The Second Variance Decomposition}
We now decompose the variance of each message propagation path into the sum over variances of its weight propagation paths. A weight propagation path of length $l$ multiplies each element within weight matrices from the input layer to the $l$-th layer. Weight propagation paths are motivated by the forward propagation of MLP: There are many paths from neurons within the input layer to a certain neuron within the $l$-th layer, each of which propagates information through layers. 

In \cref{eq:mpath}, $\delta(\cdot)$ indicates if the path $p$ is activated by ReLU, the multiplication of degree terms is a re-scaling constant, and $\vh_{j_l}^0\prod_{k_3=0}^{l-1}\mW^{k_3}$ is a FNN with $\vh_{j_l}^0$ as input. Moreover, \cref{eq:mpath} can be viewed as an FNN with activation function ReLU. As proposed by~\cite{loss_surface,poor,jknet,influence}, we can decompose such FNN into a summation over weight propagation paths from the input layer to the $l$-th layer. To see this more formally, we first provide the expression of $[h_i^l]_{p,t}$ according to~\cref{eq:mpath}. Let $\sZ(m)$ be a set $\{0, 1, \cdots, m\}$. We denote the set $\sZ(m_1^{(0)})\times \sZ(m_1^{(1)}) \cdots \sZ(m_1^{(l-1)})$ by $\hat \sZ(l-1)$. Then we have $[h_i^l]_{p,t}$ equal to:
\begin{small}
\begin{equation}
\begin{aligned}
    [h_i^l]_{p,t}=&
    (\prod_{k_1=0}^{l-1}\delta(\hat h_{j_{k_1},t}^{l-k_1-1}))(\prod_{k_2=0}^{l-1}d_{j_{k_2}j_{k_2+1}})\\
    &(\sum_{(t_0 \cdots t_{l-1}) \in \atop \hat \sZ(l-1)} h_{j_l, t_0}^0\prod_{k_3=0}^{l-1}w_{t_{k_3}, t_{k_3+1}}^k).
    \label{eq:mpaths}
\end{aligned}
\end{equation}
\end{small}

$\delta(\hat h_{j_{k_1},t}^{l-k_1-1})$ in~\cref{eq:mpaths} is the $t_{th}$ element of $\delta(\hat \vh_{j_{k_1}}^{l-k_1-1})$ in~\cref{eq:mpath}, and the $\sum_{t_0 \cdots t_{l-1}} h_{j_l, t_0}^0\prod_{k_3}w_{t_{k_3}, t_{k_3+1}}^k$ in~\cref{eq:mpaths} is the $t_{th}$ output element of the FNN $\vh_{j_l}^0\prod_{k_3=0}^{l-1}\mW^{k_3}$ in~\cref{eq:mpath}.To simplify~\cref{eq:mpaths}, let $\delta_p$ be $\prod_{k_1}\delta(\hat h_{j_{k_1},t}^{l-k_1-1})$, and $d_p$ be $\prod_{k_2}d_{j_{k_2}j_{k_2+1}}$. An output element of the FNN can be viewed as summation over weight propagation paths, where each weight propagation path, denoted as $[h_i^l]_{p,t,\phi}$, is:
\begin{equation}
[h_i^l]_{p,t,\phi}=h_{j_l, t_0}^0 w_{t_0, t_1}^0 \cdots w_{t_{l-1}, t}^{l-1},
\label{eq:wpath}
\end{equation}

where $\phi$ equals to $\{t_0\cdots t_{l-1}, t\}$ is an ordered sequence describing neurons along that weight propagation path. We denote the set including all such $\phi$ as $\Phi_{i,p,t}$. $[h_i^l]_{p,t,\phi}$ multiplies elements of input features and weight matrices across $l$ layers.  
It is obvious that $var([h_i^l]_{p,t})$ is equal to $d_p^2 var(\delta_p \sum_{\phi \in \Phi_{i,p,t}} [h_i^l]_{p,t,\phi})$. For this formula, we intend to extract $\delta_p$ from it then convert the variance of sum into the sum over variance of $[h_i^l]_{p,t,\phi}$, so that variance of each message propagation path can be expressed by the variance of weight propagation paths. To achieve this conversion, we hold the following three assumptions:

\begin{assumption}
Prior to model training, let $\phi_1$ and $\phi_2$ be two different elements of $\Phi_{i,p,t}$. Following assumptions proposed by~\cite{kaiming}, we assume the Pearson correlation between $\delta_p[h_i^l]_{p,t,\phi_1}$ and $\delta_p[h_i^l]_{p,t,\phi_2}$ is approximately 0.
\label{assu:wpaths}
\end{assumption}

\begin{assumption}
Prior to model training, the Pearson correlation between $\delta_p$ and $[h_i^l]_{p,t,\phi}$ is approximately 0, and the Pearson correlation between $\delta_p^2$ and $[h_i^l]_{p,t,\phi}^2$ is also approximately 0.
\label{assu:deltawpath}
\end{assumption}

\begin{assumption}~\cite{jknet,loss_surface,poor} assume $\delta_p$ to be a Bernoulli random variable, and different $\delta_p$ with the same length of $p$ have the same success probability. Inspired by these given assumptions, we assume such success probability to be $0.5^l$ before the model training, where $l$ is the length of $p$. 
\label{assu:delta}
\end{assumption}

When~\cref{assu:deltawpath,assu:wpaths,assu:delta} hold, $var(\delta_p \sum_{\phi \in \Phi_{i,p,t}})$ can be converted into $0.5^l var([h_i^l]_{p,t,\phi})$. According to~\cref{eq:wpath}, we are able to express $var(h_i^l)$ in terms of $var(w^k)$. Analogously, $var(\frac{\partial Loss}{\partial h_i^l})$ can also be represented in terms of $var(w^k)$. Specifically, the expressions of variance are provided as follows:

\begin{theorem}
\label{the:var_for}
Given an $L$-layer GNN defined by~\cref{eq:gcn_for}, we define $\widetilde \mA$ as a matrix in the sense that $\widetilde \mA_{ij}$ is equal to $d_{ij}$ if node $i$ and $j$ are connected, and 0 otherwise. Let $\vh^0$ be $[\mathcal{M}(\vh_0^0) \cdots \mathcal{M}(\vh_{|\sN|-1}^0)]^T$, where $\mathcal{M}(\vv)$ denotes the mean over elements of the vector $\vv$. Also, $[\vv]_i^2$ denotes the square of the $i_{th}$ element of the vector $\vv$. Then if~\cref{assu:1,assu:delta,assu:deltawpath,assu:wpaths} hold,  $var(h_i^l)$ is equal to:
\begin{small}
\begin{equation}
    var(h_i^l)=(\frac{\prod_{k_1=0}^{l-1} m_1^{(k_1)}}{2^{l}})(\prod_{k_2=0}^{l-1} var(w^{k_2})) ([\widetilde \mA^l\vh^0]^2_i).
\end{equation}
\end{small}
\end{theorem}
Next, to derive the formula of $var(\frac{\partial Loss}{\partial h_i^l})$, we need one additional lemma and assumption.
\begin{lemma} 
Let $y(i)$ denote the label of node $i$ and $s(\vv)_i$ denote the $i_{th}$ element of the softmax of the vector $\vv$. Then given a cross-entropy loss as in~\cref{subsec:gcns}, we have:
\begin{equation}
    \frac{\partial Loss}{\partial h_{i, t}^L}=
    \begin{cases}
    \frac{s(\vh_i^L)_{t}-1}{|\sN|}& t=y(i)\\
    \frac{s(\vh_i^L)_{t}}{|\sN|}& t \ne y(i).
    \end{cases}
\end{equation}
\label{lemma:1}
\vspace{-0.3cm}
\end{lemma}

\begin{assumption}
    Prior to model training, we assume the random uniformity of predicted labels at initialization and thereby $s(\vh_i^L)_{t}$ in~\cref{lemma:1} is equal to $1/C$, where $C$ is the output dimension of the last GNN layer.
\label{assu:1c}
\end{assumption}
The expression of $var(\frac{\partial Loss}{\partial h_i^l})$ is presented as the following theorem.
\begin{theorem}
\label{the:var_back}
Let $\bf{1} \in \sR^{|\sN|}$ be the vector with all 1s. Assuming the same conditions as in~\cref{the:var_for}, and the additional~\cref{assu:1c},  $var(\frac{\partial Loss}{\partial h_i^l})$ is equal to:
\begin{small}
\begin{equation}
\begin{aligned}
    var(\frac{\partial Loss}{\partial h_i^l})=&(\frac{\prod_{k_1=l+1}^{L-1}m_2^{(k_1)}(C-1)}{2^{L-l}|\sN|^2C})\\
    &(\prod_{k_2=l+1}^{L-1} var(w^{k_2}))([\widetilde \mA^{L-l}\bf{1}]^2_i).
\end{aligned}
\end{equation}
\end{small}
\end{theorem}

From \cref{the:var_for,the:var_back}, we observe that nodes at each layer have different variance since nodes have different receptive fields expressed by $[\widetilde \mA^{L-l}\bf{1}]^2_i$. This finding breaks the assumption of classic methods that all neurons at each layer have the same variance. Furthermore, while classic methods only consider the impact of hidden dimension and activation function on variance, we can see that variance is also affected by the graph structure, the message propagation of GNNs, input features and the number of nodes. Specifically, in formulas of both theorems, the constant $2$ is computed based on the \textit{ReLU} activation function following~\cite{kaiming}, the constants $m_1^{(k_1)}$, $m_2^{(k_1)}$, $C$ are input or output dimensions of weight matrices, $|\sN|$ is the number of nodes, $\widetilde \mA$ is the renormalized adjacent matrix, which is determined by the graph structure as well as the message propagation mechanism of the GNN, and $\vh^0$ is determined by input features of the given graph. We now apply these insights towards the development of a new initialization method.

\section{Proposed Virgo Initialization}

To stabilize variance for GNNs, we propose a initialization method named Virgo which incorporates the factors mentioned in the last section. The target of Virgo is to make $\sum_i var(h_i^l)$ equal to $\sum_i var(h_i^{l+1})$, and $\sum_i var(\frac{\partial Loss}{\partial h_i^l})$ equal to $\sum_i var(\frac{\partial Loss}{\partial h_i^{l+1}})$. Consideration of these two conditions then leads to the following two theorems:

\begin{theorem}
\label{the:forward}
Assuming the same conditions as in \cref{the:var_for}, to make $\sum_i var(h_i^l)$ equal to $\sum_i var(h_i^{l+1})$, we require that: 
\begin{align}
    var(w^l)=\frac{2}{m_1^{(l)}} \frac{\bf{1}^T[\widetilde \mA^{l-1}\vh^0]^2}{\bf{1}^T [\widetilde \mA^l\vh^0]^2}.
\end{align}
\end{theorem}

\begin{theorem}
\label{the:backward}
Assuming the same conditions as in~\cref{the:var_back}, to make $\sum_i var(\frac{\partial Loss}{\partial h_i^l})$ equal to $\sum_i var(\frac{\partial Loss}{\partial h_i^{l+1}})$, we require that:
\begin{align}
    var(w^l)=\frac{2}{m_2^{(l)}}\frac{\bf{1}^T[\widetilde \mA^{L-l-1}\bf{1}]^2}{\bf{1}^T[\widetilde \mA^{L-l}\bf{1}]^2}.
\end{align}
\end{theorem}
$var(w^l)$ as calculated by \cref{the:forward,the:backward} stabilizes forward and backward variances respectively. Within the variance expressions of these two theorems, the constants $2$, $m_1^{(k_1)}$, and $m_2^{(k_1)}$ are introduced to mitigate the impact of the activation function and hidden dimension on variance, analogous to factors considered by classic methods. In constrast, the appearance of $\widetilde \mA$ and $\vh^0$ encapsulate the innovative part of Virgo, which is taking graph structure, message passing and input features into account to better stabilize the variance of GNNs.

\section{Experiments}
\label{sec:exp}
In this section, we conduct experiments to investigate the performance of Virgo. Firstly, we evaluate several models on three popular graph tasks, node classification (\cref{sec:nc}), link prediction (\cref{sec:lp}) and graph classification (\cref{sec:gc}), to showcase the performance and generalizability of \method. Then to provide further insights, we test the variance stability (\cref{sec:var_sta}) of models trained with \method. In experiments below, we conduct hyperparameter sweep to search for best hyperparameter settings. To be specific, for each hyperparameter setting, we calculate the mean and standard deviation of 10 trials across different random seeds. We iterate over multiple hyperparameter settings and search for the setting with the best mean on validation datasets. We then report the mean and standard deviation on testing datasets with the selected setting as the final results. All experiments are conducted on a single Tesla T4 GPU with 16GB memory. Details of experimental setting are presented in~\cref{app_exp}

\textbf{Baseline Initializations}\ \  We compare \method\ with (i) \textit{Lecun} initialization, designed for stabilizing the forward variance of linear FNNs; (ii) \textit{Xavier} initialization, for stabilizing both forward and backward variances of linear FNNs; and (iii) \textit{Kaiming} initialization, which proposes two methods that stabilize either the forward or backward variance of CNNs activated by ReLU. We denote the two
\textit{Kaiming} variants as KaiFor and KaiBack, respectively. Similarly, we denote model initialization following \cref{the:forward} as VirgoFor, and \cref{the:backward} as VirgoBack, which stabilizes forward and backward variance respectively.

\textbf{GNN Architectures}\ \  We evaluate the model performance on node classification, link prediction and graph classification tasks with some classic GNN architectures: (i) GCN~\cite{gcn}, a spectral-based GNN for semi-supervised node classification tasks; (ii) GraphSAGE~\cite{graphsage}, stacking spatial-based convolutions to propagate message over graphs; (iii) GIN~\cite{gin}, mitigating the incapability of GNNs to distinguish different graphs structures; (iv) NGNN~\cite{ngnn} variants of (i) and (ii), which deepens GNN models with additional MLP layers interspersed with graph propagation. Note that with NGNN variants, we are able to investigate the capability of Virgo initializations on deeper GNNs more fairly while largely avoiding the effects of over-smoothing and over-squashing on model performance, i.e., Virgo is not presently designed to alleviate oversmoothing or oversquashing on deep GNNs. In practice, \method\ is directly applied to GNN layers of NGNN. All models are implemented with DGL~\cite{dgl} and PyG~\cite{pyg}.

\textbf{Datasets}\ \  For node classification, we choose three citation network datasets~\cite{cora_dataset}: cora, citeseer, pubmed, and three OGB~\cite{ogb} datasets: ogbn-arxiv, ogbn-proteins and ogbn-products. For link prediction, we adopt four OGB datasets: ogbl-ddi, ogbl-collab, ogbl-citation2 and ogbl-ppa. For graph classification, we take three social network datasests imdb\_b, imdb\_m and collab from~\cite{imdb_dataset}, and two OGB datasets ogbg-molhiv and ogbg-molpcba.

\subsection{Node Classification}
\label{sec:nc}

\textbf{Experimental setting}\ \  We use DGL to implement GCN on cora, citeseer and pubmed, and take implementations of the OGB team on the OGB node classification leaderboard to implement GCN on ogbn-arxiv and ogbn-proteins, and GraphSAGE on ogbn-products. We use neighbor sampling to support mini-batch training of GraphSAGE on ogbn-products. We tune hyper-parameters as specified previously and compare model performance of GCN and GraphSAGE with different initialization methods, including \textit{Xavier}, \textit{Lecun}, KaiFor, KaiBack, VirgoFor, and VirgoBack. We use the Adam~\cite{Adam} optimizer to update trainable parameters and use early-stop mechanism to reduce the training time overhead. 

\textbf{Results}\ \  The evaluation results are presented in \cref{tab:nc}. We observe that models with \method\ outperform models with other initializations on 5 out of 6 datasets, the lone exception being pubmed. And even on pubmed, VirgoFor and VirgoBack obtain the second and third best performance among all initializations. Furthermore, models with \method\ perform well on larger size graphs (ogbn-proteins and ogbn-products). For example, GCN initialized by \method\ on ogbn-proteins produces the largest performance gain; specifically, VirgoBack has 1.04\% higher accuracy than the best baseline initialization method \textit{Xavier}.
\begin{table*}[t]
\centering
\caption{The performance of GCN on cora, citeseer, pubmed, ogbn-arxiv and ogbn-proteins, GraphSAGE with neighbor sampling on ogbn-products. The numbers indicating the first and second place of mean accuracy are highlighted in \textcolor{red}{red} and \textcolor{blue}{blue} respectively.}
\label{tab:nc}
\vskip 0.1in

\begin{tabular}{ccccccc}
\toprule 
\multirow{2}{*}{Methods} & \multicolumn{6}{c}{Datasets} \\ 
& cora & citeseer & pubmed & arxiv & proteins & products$_{sage}$\\
\midrule
\midrule
KaiFor & 81.33±0.46 & 70.14±0.47 & 78.92±0.40 & 71.78±0.44 & 73.51±0.30 & 78.80±0.28\\
KaiBack & 81.57±0.43 & 70.79±0.49 & 79.20±0.55 & 71.44±0.37 & 73.41±0.58 & 78.73±0.24\\
Lecun & 81.41±0.33 & 70.97±0.43 & \textcolor{red}{79.48±0.31} & 71.82±0.24 & 73.29±0.44 & 78.49±0.37\\
Xavier & 81.50±0.20 & 71.09±0.52 & 79.10±0.37 & 71.74±0.32 & 73.54±0.67 & 78.89±0.31\\
\midrule
VirgoFor & \textcolor{red}{82.14±0.52} & \textcolor{red}{71.96±0.47} & \textcolor{blue}{79.42±0.42} & \textcolor{red}{72.22±0.17} & \textcolor{blue}{74.41±0.43} & \textcolor{red}{79.50±0.36}\\
VirgoBack & \textcolor{blue}{82.14±0.48} & \textcolor{blue}{71.36±0.50} & 79.34±0.22 & \textcolor{blue}{72.18±0.34} & \textcolor{red}{74.58±0.53}& \textcolor{blue}{79.45±0.36}\\
\bottomrule
\end{tabular}
\end{table*}


\subsection{Link Prediction}

\textbf{Experimental setting}\ \  We adopt the implementations of the OGB team on the OGB link prediction leaderboard for GCN and GraphSAGE, and use DGL to implement their NGNN variants. We take \textit{Xavier} and \textit{Lecun} as baseline initializations. Baselines on the leaderboard take \textit{Xavier} or \textit{Lecun} to initialize models. We observe that the reported numbers of many leaderboard submissions of baselines are significantly lower than model performance with \method, which we believe is in part an artifact of the fact that baselines on the OGB leaderboard may not be sufficiently trained. For example, GCN on ogbl-ddi achieves 37.07 hit@20 reported on leaderboard, surprisingly worse than GCN with Virgo (67.98 and 74.83). We observe that the number of training epochs of leaderboard submissions is too small to allow model training to converge. Their validation accuracy curves are still rising rather than staying flat until the end of the model training. Therefore, we re-train baselines on link prediction datasets with more epochs (for example, we take around 1000 to 2000 epochs for models on ogbl-ddi compared to 80 to 400 epochs taken by leaderboard submissions) and carefully tune their hyperparameters for a fair comparison with \method. The hyperparameter tuning setting is the same for baselines and \method. 

The evaluation metrics for models on ogbl-ddi, ogbl-collab, ogbl-citation2 and ogbl-ppa are hits@20, hits@50, mrr and hits@100, respectively. We use Adam optimizer to train models and pick the model checkpoint which has the best performance on the validation dataset. The selected checkpoint is then used to evaluate on the test dataset.

\textbf{Results}\ \  The results are presented in \cref{tab:lp}. We observe that \method\ leads to better model performance relative to other initializations in most cases. For example, NGNN-GraphSAGE with \method\ (the best one is VirgoFor with performance 80.36\%) exhibits the largest performance improvement (7.31\%) relative to NGNN-GraphSAGE with baseline initializations (the best one is \textit{Xavier} with performance 73.07\%). Overall the model performance with \method\ is best in 14 out of 16 cases. In other cases where \method is not the top 1 (NGNN-GCN and NGNN-GraphSAGE on ogbl-ppa), \method still achieves second place. Furthermore, we see that Virgo improves performance of NGNN variants, which have around 3 GCN/GraphSAGE layers and 2 MLP layers, in most cases, indicating that Virgo can be helpful to deep GNNs.
\label{sec:lp}

\begin{table*}[t]
\centering
\caption{The performance of GCN, GraphSAGE and their NGNN variants on link prediction tasks. The numbers indicating the first and second place of the average of evaluation metrics are highlighted in \textcolor{red}{red} and \textcolor{blue}{blue} respectively.}
\label{tab:lp}
\vskip 0.1in
\begin{tabular}{ccllll}
\toprule 
\multirow{2}{*}{Models} & \multirow{2}{*}{Methods} & \multicolumn{4}{c}{Datasets} \\ 
& & \makecell[c]{ddi} & \makecell[c]{collab} & \makecell[c]{citation2} & \makecell[c]{ppa}\\
\midrule
\midrule
\multirow{4}{*}{GCN} 
& Lecun & \textcolor{blue}{69.77±10.62} & 52.23±0.34 & 68.04±2.29 & 39.35±1.29\\
& Xavier & 55.16±10.64 & 53.64±0.25 & 80.88±0.18 & 37.40±0.66\\
& VirgoFor & 67.98±11.91 & \textcolor{red}{54.58±0.51} & \textcolor{blue}{81.05±0.16} & \textcolor{blue}{39.38±1.03}\\
& VirgoBack & \textcolor{red}{74.83±10.49} & \textcolor{blue}{54.31±0.42} & \textcolor{red}{81.12±0.23} & \textcolor{red}{39.85±0.89}\\
\midrule
\multirowcell{4}{NGNN-\\GCN}
& Lecun & 58.18±10.21 & 51.93±0.63 & 54.38±2.71 & 44.26±3.48\\
& Xavier & \textcolor{blue}{67.71±11.90} & 52.97±0.50 & 81.07±0.12 & \textcolor{blue}{45.47±1.64}\\
& VirgoFor & \textcolor{red}{69.05±9.54} & \textcolor{red}{54.16±0.51} & \textcolor{red}{81.41±0.28} & 45.10±1.54\\
& VirgoBack & 65.32±8.78 & \textcolor{blue}{54.13±0.38} & \textcolor{blue}{81.36±0.23} & \textcolor{red}{46.99±0.54}\\
\midrule
\multirow{4}{*}{GraphSAGE} 
& Lecun & 71.26±13.79 & 53.62±0.44 & 82.62±0.12 & \textcolor{blue}{42.98±2.38}\\
& Xavier & 70.86±10.94 & 53.48±0.34 & 83.39±0.11 & 41.99±2.42\\
& VirgoFor & \textcolor{red}{72.73±7.58} & \textcolor{blue}{53.67±0.74} & \textcolor{red}{83.74±0.01} & 42.45±0.66\\
& VirgoBack & \textcolor{blue}{72.48±9.50} & \textcolor{red}{54.16±0.47} & \textcolor{blue}{83.49±0.14} & \textcolor{red}{43.02±0.56}\\
\midrule
\multirowcell{4}{NGNN-\\GraphSAGE}
& Lecun & 65.77±13.19 & 52.14±0.43 & 81.61±0.07 & 42.41±1.48\\
& Xavier & 73.07±8.57 & 53.59±0.38 & \textcolor{red}{83.44±0.07} & \textcolor{red}{44.36±1.26}\\
& VirgoFor & \textcolor{red}{80.36±4.35} & \textcolor{red}{54.37±0.24} & 83.13±0.13 & \textcolor{blue}{44.09±0.06}\\
& VirgoBack & \textcolor{blue}{76.02±10.21} & \textcolor{blue}{53.87±0.19} & \textcolor{blue}{83.36±0.12} & 43.95±1.69\\
\bottomrule
\end{tabular}
\end{table*}

\subsection{Graph Classification}
\label{sec:gc}

\textbf{Experimental setting}\ \  We use DGL to implement models on imdb\_b, imdb\_m and collab, and we take implementations of the OGB team on the OGB graph classification leaderboard to conduct experiments on OGB datasets. We evaluate GCN and GIN with \textit{Xavier}, \textit{Lecun} and \method. To compute the initial variance of weight matrices based on \method, we sample a subset of training graphs into a single graph defined as the approximation graph, and utilize its graph structure to approximate $\widetilde \mA$ in \cref{the:forward,the:backward}. For imdb\_b, imdb\_m, collab and ogbg-molhiv, we merge all training graphs into the approximation graph. For ogbg-molpcba, we randomly and uniformly sample 10,000 training graphs and merge them into the approximation graph since its training dataset is too large to be fed into a single GPU. We use Adam optimizer to update trainable parameters. For OGB datasets, we take the model checkpoint that has the best validation performance, and evaluate this checkpoint on test datasets to obtain the test performance. We evaluate classification accuracy on imdb and collab, roc-auc on ogbg-molhiv, and average precision on ogbg-molpcba.

\textbf{Results}\ \  The experimental results are reported in \cref{tab:gc}. We have three observations: First, both VirgoFor and VirgoBack take top 2 in 8 out of 10 cases, and 9 out of 10 cases have \method\ as their best initialization method. These results shows the benefit of \method\ to model performance, which is not limited to node classification and link prediction tasks. In other cases where \method\ does not occupy top 2 positions (GCN on imdb\_b and ogbg-molhiv), it takes at least the second place. Second, by comparing the best model performance with \method\ and the best one with baseline initializations, \method\ brings at least 1\% improvements to GCN on 4 out of 5 datasets: imdb\_b (1.23\%), collab (1.53\%), ogbg-molhiv (1.72\%) and ogbg-molpcba (1.11\%). Finally, performance improvements on ogbg-molpcba with trivial sampling methods indicate that we can simply use random uniform sampling methods as described previously to achieve competitive performance with Virgo.

\begin{table*}[t]
\caption{The performance of GCN and GIN on graph classification tasks. The numbers indicating the first and second place of the average of evaluation metrics are highlighted in \textcolor{red}{red} and \textcolor{blue}{blue} respectively.}
\label{tab:gc}
\centering
\begin{tabular}{ccccccc}
\toprule 
\multirow{2}{*}{Models} & \multirow{2}{*}{Methods} & \multicolumn{5}{c}{Datasets} \\ 
& & imdb$_b$ & imdb$_m$ & collab & molhiv & molpcba\\
\midrule
\midrule
\multirow{3}{*}{GCN} 
& Lecun & 74.57±2.19 & 52.04±1.77 & 82.43±0.84 & 77.06±0.63 & 20.44±0.21 \\
& Xavier & 74.17±2.12 & 51.83±3.23 & 82.17±1.13 & 76.49±0.95 & 20.62±0.31 \\
& VirgoFor & \textcolor{red}{75.80±2.32} & \textcolor{blue}{52.13±2.00} & \textcolor{red}{83.96±0.78} & \textcolor{blue}{77.94±0.58} & \textcolor{blue}{21.12±0.50}\\
& VirgoBack & \textcolor{blue}{75.40±4.84} & \textcolor{red}{52.67±2.15} & \textcolor{blue}{83.44±0.73} & \textcolor{red}{78.78±0.16} & \textcolor{red}{21.73±0.15}\\
\midrule
\multirow{3}{*}{GIN} 
& Lecun & 75.07±2.43 & 52.61±1.41 & 83.10±0.74 & \textcolor{blue}{77.12±1.01} & 23.45±0.23 \\ 
& Xavier & \textcolor{red}{75.42±3.31} & 52.24±1.57 & 83.27±1.39 & 76.16±1.76 & 23.01±0.33\\
& VirgoFor & \textcolor{blue}{75.20±3.66} & \textcolor{blue}{53.20±1.36} & \textcolor{red}{84.32±0.63} & 77.09±1.19 & \textcolor{blue}{24.15±0.49}\\
& VirgoBack & 74.60±2.73 & \textcolor{red}{53.60±2.00} & \textcolor{blue}{83.84±1.17} & \textcolor{red}{77.90±1.43} & \textcolor{red}{24.23±0.12}\\
\bottomrule
\end{tabular}
\end{table*}

\subsection{Variance Stability}
\label{sec:var_sta}

\begin{figure}[t]
\includegraphics[width=.9\columnwidth]{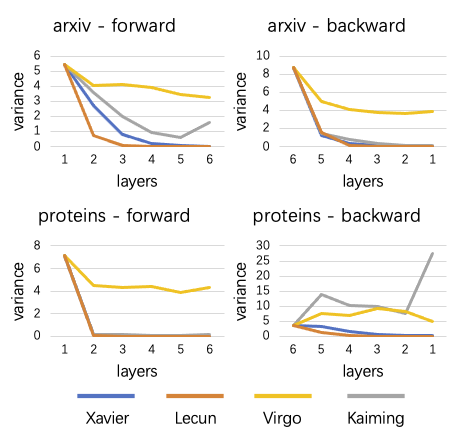}
\caption{Variance stability of GCN on ogbn-arxiv and ogbn-proteins at the initialization. \textit{Upper left}: Forward variance on ogbn-arxiv$ (\times 10^{-2})$. \textit{Upper right}: Backward variance on ogbn-arxiv$ (\times 10^{-25})$. \textit{Bottom left}: Forward variance on ogbn-proteins $(\times 10^{-3})$. \textit{Bottom right}: Backward variance on ogbn-proteins$(\times 10^{-16})$. As for Kaiming and Virgo, we adopt KaiFor and VirgoFor in figures labeled \textit{forward}, and KaiBack and VirgoBack in figures labeled \textit{backward}.}
\label{fig:var}

\end{figure}

In this section, we compare variance stability of GCN following~\cref{eq:gcn_for} on ogbn-arxiv and ogbn-proteins at initialization with different methods. For each combination of an initialization method and a dataset, we pick the best hyperparameter setting that has been investigated in~\cref{sec:nc}, and test forward and backward variance across 5 layers. Specifically, we compute the variance of each node, and compute the mean of variance over nodes at each layer. The results are presented in~\cref{fig:var}. We observe that \method\ leads to more stable variance change than classic methods. 
For example, in the cases of backward variance on ogbn-arxiv and forward variance on ogbn-proteins, only \method\ mitigates the steep decline in variances towards zero, emblematic of the importance of accounting for graph structure and message passing relative to other factors in stabilizing the variance.


\section{Related Work}
Our work is closely related to Graph Neural Networks (GNNs) and initialization methods. We introduce classic work in these fields as follows.

\paragraph{Graph Neural Networks} There are a number of approaches that generalize convolution operations on images to the graph domain and achieve state-of-the-art performance on popular tasks of graphs, such as node classification, link prediction and graph classification. ~\cite{chebnet,cayleynet,gcn} propose spectral-based convolutional neural networks based on graph Laplacian matrix for learning on graphs. ~\cite{sgc} achieves comparable peformance with GCN~\cite{gcn} while reduce excess complexity. ~\cite{graphsage} propose spatial-based convolutions to propagate message over graphs based on nodes' spatial dependencies. ~\cite{gat,gatv2} utilize attention mechanism to learn contributions of neighboring nodes to the target nodes. ~\cite{gin} investigates the expressive power of classic GNNs and develop a simple but more powerful structure. ~\cite{jknet,deepgcn} are designed to learn higher level of knowledge from graphs by increasing the number of GCN layers.  Inspired by ~\cite{nin,gin}, \cite{ngnn} equips each GCN layer with multiple linear layers to form a NGNN block, which extracts more complex semantics from graphs. 


\paragraph{Initialization methods} Several initialization methods have been proposed to define initial values of model parameters prior to the model training. \textit{Lecun} initialization~\cite{lecun} requires forward variance to be 1. \textit{Xavier} initialization~\cite{xavier} is similar to \textit{Lecun} but considers both forward and backward variance. Despite \textit{Lecun} and \textit{Xavier} assuming no non-linearity in neural networks, they work well in many applications. \textit{Kaiming} initialization~\cite{kaiming} extends \textit{Xavier} to CNNs with ReLU non-linearity. ~\cite{orthogonal} exhibit a new class of orthogonal matrix initialization for deep linear neural networks. ~\cite{allyouneed} proposes \textit{LSUV} initialization based on ~\cite{orthogonal} to consider the impact of more model components on variance, such as tanh and maxout. ~\cite{rwi} designs a \textit{Random Walk} initialization for FNNs with non-linearity to keep constant the logarithm of squared magnitude of gradients acorss all layers. ~\cite{oldgold} proposes a topology-aware isometric initialization to facilitate gradient flow during GCN training.~\cite{mlpinit} initialize GNNs with pre-trained MLP parameters to train GNNs more efficiently. However, most existing work does not directly apply to GNNs because of: only analyzing output and gradients with FNNs~\cite{lecun,xavier,kaiming,orthogonal,allyouneed,rwi}, ignoring the impact of non-linearities~\cite{lecun,xavier,orthogonal,oldgold}, assuming outputs and gradients of neurons at each layer are i.i.d~\cite{lecun,xavier,kaiming}. In contrast, our analysis is explicitly based on GNNs over graphs, and accounts for the fact that the outputs and gradients of different nodes have correlated variances due to the effective receptive fields at each layer. We then exploit these findings to develop \method, which is better equipped to stabilize GNN variances.

\section{Conclusion}
In this paper, we derive explicit expressions for the forward and backward variance of GNN initializations, and analyze deficiencies of classic initialization methods when applied to stabilizing them.  Informed by this perspective, we propose a new GNN initialization scheme Virgo, and conduct comprehensive experiments to compare with 4 classic initialization methods on 15 datasets across 3 popular graph learning tasks showing superior performance.

In the future, there are two shortcomings of \method that could potentially be addressed: (i) \method\ is derived based on GNNs that have pre-computed constant coefficients between neighboring nodes, and thus it cannot directly be generalized to GNNs like GAT~\cite{gat} with adaptive/learnable coefficients between neighbors.  (ii) \method\ only considers one level of aggregation, thus it is not yet suitable for models like RGCN~\cite{rgcn} that have multiple levels of aggregation.  Beyond these considerations, we do not believe that our approach will have any undue negative societal impact beyond the minor potential essentially shared by all GNN methods, e.g., propagating unfair biases, etc.



\bibliography{example_paper}

\begin{thebibliography}{40}
\providecommand{\natexlab}[1]{#1}
\providecommand{\url}[1]{\texttt{#1}}
\expandafter\ifx\csname urlstyle\endcsname\relax
  \providecommand{\doi}[1]{doi: #1}\else
  \providecommand{\doi}{doi: \begingroup \urlstyle{rm}\Url}\fi

\bibitem[Brody et~al.(2021)Brody, Alon, and Yahav]{gatv2}
Brody, S., Alon, U., and Yahav, E.
\newblock How attentive are graph attention networks?
\newblock \emph{arXiv preprint arXiv:2105.14491}, 2021.

\bibitem[Choromanska et~al.(2015)Choromanska, Henaff, Mathieu, Arous, and
  LeCun]{loss_surface}
Choromanska, A., Henaff, M., Mathieu, M., Arous, G.~B., and LeCun, Y.
\newblock The loss surfaces of multilayer networks.
\newblock In \emph{Artificial intelligence and statistics}, pp.\  192--204.
  PMLR, 2015.

\bibitem[Defferrard et~al.(2016)Defferrard, Bresson, and
  Vandergheynst]{chebnet}
Defferrard, M., Bresson, X., and Vandergheynst, P.
\newblock Convolutional neural networks on graphs with fast localized spectral
  filtering.
\newblock \emph{Advances in neural information processing systems}, 29, 2016.

\bibitem[Fan et~al.(2019)Fan, Ma, Li, He, Zhao, Tang, and Yin]{fan2019graph}
Fan, W., Ma, Y., Li, Q., He, Y., Zhao, E., Tang, J., and Yin, D.
\newblock Graph neural networks for social recommendation.
\newblock In \emph{The World Wide Web Conference}, pp.\  417--426, 2019.

\bibitem[Fey \& Lenssen(2019)Fey and Lenssen]{pyg}
Fey, M. and Lenssen, J.~E.
\newblock Fast graph representation learning with pytorch geometric.
\newblock \emph{arXiv preprint arXiv:1903.02428}, 2019.

\bibitem[Gasteiger et~al.(2022)Gasteiger, Qian, and G{\"u}nnemann]{influence}
Gasteiger, J., Qian, C., and G{\"u}nnemann, S.
\newblock Influence-based mini-batching for graph neural networks.
\newblock \emph{arXiv preprint arXiv:2212.09083}, 2022.

\bibitem[Glorot \& Bengio(2010)Glorot and Bengio]{xavier}
Glorot, X. and Bengio, Y.
\newblock Understanding the difficulty of training deep feedforward neural
  networks.
\newblock In \emph{Proceedings of the thirteenth international conference on
  artificial intelligence and statistics}, pp.\  249--256. JMLR Workshop and
  Conference Proceedings, 2010.

\bibitem[Hamilton et~al.(2017)Hamilton, Ying, and Leskovec]{graphsage}
Hamilton, W.~L., Ying, R., and Leskovec, J.
\newblock Inductive representation learning on large graphs.
\newblock In \emph{Proceedings of the 31st International Conference on Neural
  Information Processing Systems}, pp.\  1025--1035, 2017.

\bibitem[Han et~al.(2022)Han, Zhao, Liu, Hu, and Shah]{mlpinit}
Han, X., Zhao, T., Liu, Y., Hu, X., and Shah, N.
\newblock Mlpinit: Embarrassingly simple gnn training acceleration with mlp
  initialization.
\newblock \emph{arXiv preprint arXiv:2210.00102}, 2022.

\bibitem[He et~al.(2015)He, Zhang, Ren, and Sun]{kaiming}
He, K., Zhang, X., Ren, S., and Sun, J.
\newblock Delving deep into rectifiers: Surpassing human-level performance on
  imagenet classification.
\newblock In \emph{Proceedings of the IEEE international conference on computer
  vision}, pp.\  1026--1034, 2015.

\bibitem[Hu et~al.(2020)Hu, Fey, Zitnik, Dong, Ren, Liu, Catasta, and
  Leskovec]{ogb}
Hu, W., Fey, M., Zitnik, M., Dong, Y., Ren, H., Liu, B., Catasta, M., and
  Leskovec, J.
\newblock Open graph benchmark: Datasets for machine learning on graphs.
\newblock \emph{arXiv preprint arXiv:2005.00687}, 2020.

\bibitem[Jaiswal et~al.(2022)Jaiswal, Wang, Chen, Rousseau, Ding, and
  Wang]{oldgold}
Jaiswal, A., Wang, P., Chen, T., Rousseau, J., Ding, Y., and Wang, Z.
\newblock Old can be gold: Better gradient flow can make vanilla-gcns great
  again.
\newblock \emph{Advances in Neural Information Processing Systems},
  35:\penalty0 7561--7574, 2022.

\bibitem[Jing \& Xu(2021)Jing and Xu]{jing2021fast}
Jing, X. and Xu, J.
\newblock Fast and effective protein model refinement using deep graph neural
  networks.
\newblock \emph{Nature Computational Science}, 1\penalty0 (7):\penalty0
  462--469, 2021.

\bibitem[Kawaguchi(2016)]{poor}
Kawaguchi, K.
\newblock Deep learning without poor local minima.
\newblock \emph{Advances in neural information processing systems}, 29, 2016.

\bibitem[Kingma \& Ba(2014)Kingma and Ba]{Adam}
Kingma, D.~P. and Ba, J.
\newblock Adam: A method for stochastic optimization.
\newblock \emph{arXiv preprint arXiv:1412.6980}, 2014.

\bibitem[Kipf \& Welling(2016)Kipf and Welling]{gcn}
Kipf, T.~N. and Welling, M.
\newblock Semi-supervised classification with graph convolutional networks.
\newblock \emph{arXiv preprint arXiv:1609.02907}, 2016.

\bibitem[LeCun et~al.(2012)LeCun, Bottou, Orr, and M{\"u}ller]{lecun}
LeCun, Y.~A., Bottou, L., Orr, G.~B., and M{\"u}ller, K.-R.
\newblock Efficient backprop.
\newblock In \emph{Neural networks: Tricks of the trade}, pp.\  9--48.
  Springer, 2012.

\bibitem[Levie et~al.(2018)Levie, Monti, Bresson, and Bronstein]{cayleynet}
Levie, R., Monti, F., Bresson, X., and Bronstein, M.~M.
\newblock Cayleynets: Graph convolutional neural networks with complex rational
  spectral filters.
\newblock \emph{IEEE Transactions on Signal Processing}, 67\penalty0
  (1):\penalty0 97--109, 2018.

\bibitem[Li et~al.(2019)Li, Muller, Thabet, and Ghanem]{deepgcn}
Li, G., Muller, M., Thabet, A., and Ghanem, B.
\newblock Deepgcns: Can gcns go as deep as cnns?
\newblock In \emph{Proceedings of the IEEE/CVF international conference on
  computer vision}, pp.\  9267--9276, 2019.

\bibitem[Lin et~al.(2013)Lin, Chen, and Yan]{nin}
Lin, M., Chen, Q., and Yan, S.
\newblock Network in network.
\newblock \emph{arXiv preprint arXiv:1312.4400}, 2013.

\bibitem[Liu et~al.(2020)Liu, Dou, Yu, Deng, and Peng]{liu2020alleviating}
Liu, Z., Dou, Y., Yu, P.~S., Deng, Y., and Peng, H.
\newblock Alleviating the inconsistency problem of applying graph neural
  network to fraud detection.
\newblock In \emph{Proceedings of the 43rd International ACM SIGIR Conference
  on Research and Development in Information Retrieval}, pp.\  1569--1572,
  2020.

\bibitem[Mishkin \& Matas(2015)Mishkin and Matas]{allyouneed}
Mishkin, D. and Matas, J.
\newblock All you need is a good init.
\newblock \emph{arXiv preprint arXiv:1511.06422}, 2015.

\bibitem[Monti et~al.(2017)Monti, Boscaini, Masci, Rodola, Svoboda, and
  Bronstein]{monet}
Monti, F., Boscaini, D., Masci, J., Rodola, E., Svoboda, J., and Bronstein,
  M.~M.
\newblock Geometric deep learning on graphs and manifolds using mixture model
  cnns.
\newblock In \emph{Proceedings of the IEEE conference on computer vision and
  pattern recognition}, pp.\  5115--5124, 2017.

\bibitem[Nguyen et~al.(2020)Nguyen, Le, Quinn, Le, and
  Venkatesh]{nguyen2020predicting}
Nguyen, T., Le, H., Quinn, T.~P., Le, T., and Venkatesh, S.
\newblock Predicting drug--target binding affinity with graph neural networks.
\newblock \emph{BioRxiv}, pp.\  684662, 2020.

\bibitem[Rossi et~al.(2020)Rossi, Chamberlain, Frasca, Eynard, Monti, and
  Bronstein]{rossi2020temporal}
Rossi, E., Chamberlain, B., Frasca, F., Eynard, D., Monti, F., and Bronstein,
  M.
\newblock Temporal graph networks for deep learning on dynamic graphs.
\newblock \emph{arXiv preprint arXiv:2006.10637}, 2020.

\bibitem[Saxe et~al.(2013)Saxe, McClelland, and Ganguli]{orthogonal}
Saxe, A.~M., McClelland, J.~L., and Ganguli, S.
\newblock Exact solutions to the nonlinear dynamics of learning in deep linear
  neural networks.
\newblock \emph{arXiv preprint arXiv:1312.6120}, 2013.

\bibitem[Schlichtkrull et~al.(2018)Schlichtkrull, Kipf, Bloem, Berg, Titov, and
  Welling]{rgcn}
Schlichtkrull, M., Kipf, T.~N., Bloem, P., Berg, R. v.~d., Titov, I., and
  Welling, M.
\newblock Modeling relational data with graph convolutional networks.
\newblock In \emph{European semantic web conference}, pp.\  593--607. Springer,
  2018.

\bibitem[Sen et~al.(2008)Sen, Namata, Bilgic, Getoor, Galligher, and
  Eliassi-Rad]{cora_dataset}
Sen, P., Namata, G., Bilgic, M., Getoor, L., Galligher, B., and Eliassi-Rad, T.
\newblock Collective classification in network data.
\newblock \emph{AI magazine}, 29\penalty0 (3):\penalty0 93--93, 2008.

\bibitem[Song et~al.(2021)Song, Ma, Li, Zhang, and Wipf]{ngnn}
Song, X., Ma, R., Li, J., Zhang, M., and Wipf, D.~P.
\newblock Network in graph neural network.
\newblock \emph{arXiv preprint arXiv:2111.11638}, 2021.

\bibitem[Strokach et~al.(2020)Strokach, Becerra, Corbi-Verge, Perez-Riba, and
  Kim]{strokach2020fast}
Strokach, A., Becerra, D., Corbi-Verge, C., Perez-Riba, A., and Kim, P.~M.
\newblock Fast and flexible protein design using deep graph neural networks.
\newblock \emph{Cell systems}, 11\penalty0 (4):\penalty0 402--411, 2020.

\bibitem[Sussillo \& Abbott(2014)Sussillo and Abbott]{rwi}
Sussillo, D. and Abbott, L.
\newblock Random walk initialization for training very deep feedforward
  networks.
\newblock \emph{arXiv preprint arXiv:1412.6558}, 2014.

\bibitem[Veli{\v{c}}kovi{\'c} et~al.(2017)Veli{\v{c}}kovi{\'c}, Cucurull,
  Casanova, Romero, Lio, and Bengio]{gat}
Veli{\v{c}}kovi{\'c}, P., Cucurull, G., Casanova, A., Romero, A., Lio, P., and
  Bengio, Y.
\newblock Graph attention networks.
\newblock \emph{arXiv preprint arXiv:1710.10903}, 2017.

\bibitem[Wang et~al.(2019)Wang, Wen, Wu, Huang, and Xion]{wang2019fdgars}
Wang, J., Wen, R., Wu, C., Huang, Y., and Xion, J.
\newblock Fdgars: Fraudster detection via graph convolutional networks in
  online app review system.
\newblock In \emph{Companion Proceedings of The 2019 World Wide Web
  Conference}, pp.\  310--316, 2019.

\bibitem[Wang et~al.(2020)Wang, Zheng, Ye, Gan, Li, Song, Zhou, Ma, Yu, Gai,
  Xiao, He, Karypis, Li, and Zhang]{dgl}
Wang, M., Zheng, D., Ye, Z., Gan, Q., Li, M., Song, X., Zhou, J., Ma, C., Yu,
  L., Gai, Y., Xiao, T., He, T., Karypis, G., Li, J., and Zhang, Z.
\newblock Deep graph library: A graph-centric, highly-performant package for
  graph neural networks, 2020.

\bibitem[Wu et~al.(2019)Wu, Souza, Zhang, Fifty, Yu, and Weinberger]{sgc}
Wu, F., Souza, A., Zhang, T., Fifty, C., Yu, T., and Weinberger, K.
\newblock Simplifying graph convolutional networks.
\newblock In \emph{International conference on machine learning}, pp.\
  6861--6871. PMLR, 2019.

\bibitem[Xu et~al.(2018{\natexlab{a}})Xu, Hu, Leskovec, and Jegelka]{gin}
Xu, K., Hu, W., Leskovec, J., and Jegelka, S.
\newblock How powerful are graph neural networks?
\newblock \emph{arXiv preprint arXiv:1810.00826}, 2018{\natexlab{a}}.

\bibitem[Xu et~al.(2018{\natexlab{b}})Xu, Li, Tian, Sonobe, Kawarabayashi, and
  Jegelka]{jknet}
Xu, K., Li, C., Tian, Y., Sonobe, T., Kawarabayashi, K., and Jegelka, S.
\newblock Representation learning on graphs with jumping knowledge networks.
\newblock \emph{CoRR}, abs/1806.03536, 2018{\natexlab{b}}.

\bibitem[Yanardag \& Vishwanathan(2015)Yanardag and Vishwanathan]{imdb_dataset}
Yanardag, P. and Vishwanathan, S.
\newblock Deep graph kernels.
\newblock In \emph{Proceedings of the 21th ACM SIGKDD international conference
  on knowledge discovery and data mining}, pp.\  1365--1374, 2015.

\bibitem[Ying et~al.(2018)Ying, He, Chen, Eksombatchai, Hamilton, and
  Leskovec]{ying2018graph}
Ying, R., He, R., Chen, K., Eksombatchai, P., Hamilton, W.~L., and Leskovec, J.
\newblock Graph convolutional neural networks for web-scale recommender
  systems.
\newblock In \emph{Proceedings of the 24th ACM SIGKDD International Conference
  on Knowledge Discovery \& Data Mining}, pp.\  974--983, 2018.

\bibitem[Yu et~al.(2021)Yu, Lin, Liu, Ge, Ou, and Qin]{yu2021self}
Yu, W., Lin, X., Liu, J., Ge, J., Ou, W., and Qin, Z.
\newblock Self-propagation graph neural network for recommendation.
\newblock \emph{IEEE Transactions on Knowledge and Data Engineering}, 2021.

\end{thebibliography}
\bibliographystyle{icml2023}

\newpage
\appendix
\onecolumn
\section{Proof and Empirical Verification}
\label{appendix}
In this section, we show empirical verification of assumptions, and proofs of one lemma and multiple theorems in the main body of this paper. All experiments are conducted on a 3-layer GCN initialized by \textit{Xavier} initialization. We argue that the initialization method does not affect empirical results in this section since they are not related to distributions of GCN weight matrices. 

\textbf{\cref{assu:1}}\ \ We first take a forward propagation of GCN on the input graph and obtain hidden embeddings of 3 layers. Then we randomly sample two nodes $u$ and $v$ from the input graph, and take all paths of length 3 from $u$ to $v$ in the graph if there exists at least one. The random sampling process continues until the number of paths reach 100. Let's recap that the expression of message propagation paths is:

\begin{equation}
[\vh_i^l]_{p}=(\prod_{k_1=0 \atop \odot}^{l-1}\delta(\hat \vh_{j_{k_1}}^{l-k_1-1}))(\prod_{k_2=0}^{l-1}d_{j_{k_2}j_{k_2+1}})(\vh_{j_l}^0\prod_{k_3=0}^{l-1}\mW^{k_3})
\label{proof:mpath}
\end{equation}

It's obvious that sampled paths are instances of $p$ that has length 3 in~\cref{proof:mpath}. For~\cref{proof:mpath}, the node $v$ and $u$ of each sampled path are the destination node $i$ and the starting node $j_l$ of the $p$, respectively. We apply $\delta$ to pre-activated embeddings of nodes to estimate $\delta(\hat \vh_{j_{k_1}}^{l-k_1-1})$, then calculate the multiplication of resulted quantities along each path to estimate $\prod_{k_1, \odot}\delta(\hat \vh_{j_{k_1}}^{l-k_1-1})$. We take the multiplication of degrees of nodes along the path $p$ as $\prod_{k_2=0}^{l-1}d_{j_{k_2}j_{k_2+1}}$, and take the multiplication of weight matrices across 3 GCN layers with input features of node $j_l$ as $\vh_{j_l}^0 \prod_{k_3=0}^{l-1}\mW^{k_3}$. As a result, we are able to estimate $[\vh_i^l]_{p}$ with quantities mentioned above. Next, we randomly sample 100 neurons from each $[\vh_i^l]_{p}$ as its 100 $[\vh_i^l]_{p,t}$, and calculate the Pearson correlation between each pair of $[\vh_i^l]_{p_1,t}$ and $[\vh_i^l]_{p_2,t}$, where $p_1$ is not equal to $p_2$. Thereby we obtain 4950(pairs) resulted numbers, which indicate the Pearson correlation between message propagation paths of length 3. We take the average and standard deviation of them, and put results in~\cref{tab:est_mpaths}.

In~\cref{tab:est_mpaths}, we show evaluation results of 3-layer GCN on cora, pubmed, citeseer and ogbn-arxiv. We test the Pearson correlation of message propagation paths that have length 1, 2 besides 3. The row titled with \textbf{Expected} indicates that we require the Pearson correlation to be 1 to hold~\cref{assu:1}. We can see that the Pearson correlation on all datasets are greater than 0.83, especially on ogbn-arxiv where results are greater than 0.9.

\begin{table}[htbp]
\vspace{-0.3cm}
 \centering
 \caption{Estimation of the Pearson correlation between different message propagation paths.} 
 \vskip 0.1in
 \label{tab:est_mpaths}
\begin{tabular}{ccccccc}
\toprule 
\multirow{2}{*}{Dataset} & \multicolumn{3}{c}{Layers} \\ & 1 & 2 & 3\\
\midrule
\midrule
cora & 0.83$\pm$0.13 & 0.86$\pm$0.12 & 0.84$\pm$0.15\\
\midrule
pubmed & 0.89$\pm$0.09 & 0.89$\pm$0.09 & 0.93$\pm$0.06 \\
\midrule
citeseer & 0.83$\pm$0.12 & 0.85$\pm$0.11 & 0.85$\pm$0.12 \\
\midrule
arxiv & 0.91$\pm$0.06 & 0.91$\pm$0.05 & 0.94$\pm$0.04 \\
\midrule
Expected & 1.0 & 1.0 & 1.0\\
\bottomrule
\end{tabular}
\end{table}

\textbf{\cref{assu:wpaths}}\ \ We use the same experiment setting as in the empirical verification of~\cref{assu:1}. For each $\delta_p[\vh_i^l]_{p,t}$, we randomly take 50 neurons as observations of $\delta_p[h_i^l]_{p,t,\phi_1}$ and take remain 50 neurons as observations of $\delta_p[h_i^l]_{p,t,\phi_2}$. We have 4950(pairs) of $\delta_p[h_i^l]_{p,t,\phi_1}$ and $\delta_p[h_i^l]_{p,t,\phi_2}$, then we calculate the Pearson correlation between each pair. We take the average and standard deviation of resulted numbers and report results in~\cref{tab:est_wpaths}. 

In~\cref{tab:est_wpaths}, the row titled with \textbf{Expected} indicates that we require the Pearson correlation to be 0 to hold~\cref{assu:wpaths}. We observe that all average numbers of the Pearson correlation are less than 0.1.

\begin{table}[htbp]
\vspace{-0.3cm}
 \centering
 \caption{Estimation of the Pearson correlation between different weight propagation paths.} 
 \label{tab:est_wpaths}
 \vskip 0.1in
\begin{tabular}{ccccccc}
\toprule 
\multirow{2}{*}{Dataset} & \multicolumn{3}{c}{Layers} \\ & 1 & 2 & 3\\
\midrule
\midrule
cora & 0.07$\pm$0.05 & 0.07$\pm$0.07 & 0.05$\pm$0.07\\
\midrule
pubmed & 0.06$\pm$0.04 & 0.07$\pm$0.06 & 0.04$\pm$0.06 \\
\midrule
citeseer &  0.07$\pm$0.05 & 0.06$\pm$0.05 & 0.03$\pm$0.04 \\
\midrule
arxiv & 0.07$\pm$0.06 & 0.08$\pm$0.07 & 0.07$\pm$0.06 \\
\midrule
Expected & 0.0 & 0.0 & 0.0\\
\bottomrule
\end{tabular}
\end{table}

\textbf{\cref{assu:deltawpath}}\ \ We use the same experiment setting as in the empirical verification of~\cref{assu:1}. Let's recap expressions of $\delta_p$ and $[h_i^l]_{p,t,\phi}$:
\begin{align}
    \delta_p&=\prod_{k=0}^{l-1}\delta(\hat h_{j_{k},t}^{l-k-1})\\
    [h_i^l]_{p,t,\phi}&=h_{j_l, t_0}^0 w_{t_0, t_1}^0 \cdots w_{t_{l-1}, t}^{l-1}
\end{align}

For each message propagation path $p$, we take the estimation of neurons of $\prod_{k_1, \odot}\delta(\hat \vh_{j_{k_1}}^{l-k_1-1})$ in the empirical verification of~\cref{assu:1} to estimate $\delta_p$ and take the estimation of neurons of $\vh_{j_l}^0 \prod_{k_3=0}^{l-1}\mW^{k_3}$ in the empirical verification of~\cref{assu:1} to estimate $[h_i^l]_{p,t,\phi}$. In practice, we will obtain 100(message propagation paths) pairs of both $\delta_p$ and $[h_i^l]_{p,t,\phi}$. We then report the average and standard deviation of Pearson correlation between $\delta_p$ and $[h_i^l]_{p,t,\phi}$, between $\delta_p^2$ and $[h_i^l]^2_{p,t,\phi}$, in~\cref{tab:est_delta_wpaths}.

In~\cref{tab:est_delta_wpaths}, the row titled with \textbf{Expected} indicates that we require the Pearson correlation to be 0 to hold~\cref{assu:deltawpath}. We observe that all average numbers of the Pearson correlation are less than or equal to 0.1. 

\begin{table}[htbp]
\vspace{-0.45cm}
 \centering
 \caption{Estimation of the Pearson correlation between $\delta_p$ and $[h_i^l]_{p,t,\phi}$, between $\delta_p^2$ and $[h_i^l]^2_{p,t,\phi}$. We present the the Pearson correlation between $\delta_p$ and $[h_i^l]_{p,t,\phi}$ in rows that have dataset names with subscripts 1, and the Pearson correlation between $\delta_p^2$ and $[h_i^l]^2_{p,t,\phi}$ in rows that have dataset names with subscripts 2.} 
 \label{tab:est_delta_wpaths}
 \vskip 0.1in
\begin{tabular}{cccccc}
\toprule 
\multirow{2}{*}{Dataset} & \multicolumn{3}{c}{Layers} \\ & 1 & 2 & 3\\
\midrule
\midrule
cora$_1$ & 0.04$\pm$0.03 & 0.05$\pm$0.03 & 0.06$\pm$0.04\\
cora$_2$ & 0.05$\pm$0.03 & 0.04$\pm$0.03 & 0.03$\pm$0.03\\
\midrule
pubmed$_1$ & 0.04$\pm$0.03 & 0.04$\pm$0.03 & 0.06$\pm$0.04 \\
pubmed$_2$ & 0.05$\pm$0.04 & 0.04$\pm$0.03 & 0.03$\pm$0.03 \\
\midrule
citeseer$_1$ & 0.04$\pm$0.03 & 0.05$\pm$0.04 & 0.04$\pm$0.03 \\
citeseer$_2$ & 0.04$\pm$0.05 & 0.05$\pm$0.03 & 0.05$\pm$0.03\\
\midrule
arxiv$_1$ & 0.02$\pm$0.01 & 0.03$\pm$0.01 & 0.1$\pm$0.01 \\
arxiv$_2$ & 0.03$\pm$0.01 & 0.04$\pm$0.01 & 0.01$\pm$0.01\\
\midrule
Expected & 0.0 & 0.0 & 0.0\\
\bottomrule
\end{tabular}
\end{table}

\textbf{\cref{assu:delta}}\ \ Assuming that $\delta_p$ is a Bernoulli random variable, it's known that mean of a Bernoulli random variable is equal to its success probability. With the same experiment setting as in the empirical verification of~\cref{assu:deltawpath}, we have 100(neurons) * 100(message propagation paths) observations of $\delta_p$. We calculate the mean values and standard deviation of $\delta_p$ and report results in~\cref{tab:est_delta} 

In~\cref{tab:est_delta}, the row titled with \textbf{Expected} indicates that we require the mean of $\delta_p$ to be 0.5 to hold~\cref{assu:delta}. We observe that averages of $\delta_p$ are quite close to 0.5.

\begin{table}[htbp]
\vspace{-0.3cm}
 \centering
 \caption{Estimation of the success probability of $\delta_p$.} 
 \label{tab:est_delta}
 \vskip 0.1in
\begin{tabular}{ccccccc}
\toprule 
\multirow{2}{*}{Dataset} & \multicolumn{3}{c}{Layers} \\ & 1 & 2 & 3\\
\midrule
\midrule
cora & 0.47$\pm$0.03 & 0.24$\pm$0.02 & 0.13$\pm$0.01\\
\midrule
pubmed & 0.50$\pm$0.03 & 0.25$\pm$0.02 & 0.13$\pm$0.02 \\
\midrule
citeseer &  0.52$\pm$0.03 & 0.27$\pm$0.02 & 0.14$\pm$0.02\\
\midrule
arxiv & 0.54$\pm$0.01 & 0.27$\pm$0.01 & 0.11$\pm$0.01 \\
\midrule
Expected & 0.5 & 0.25 & 0.125\\
\bottomrule
\end{tabular}
\end{table}

\textbf{\cref{the:var_for}}\ \ 
Let's recap the forward propagation of the neuron $t$ of the node $i$ at the path $p$ is presented as follows:

\begin{equation}
    [h_i^l]_{p,t}=
    (\prod_{k_1=0}^{l-1}\delta(\hat h_{j_{k_1},t}^{l-k_1-1}))(\prod_{k_2=0}^{l-1}d_{j_{k_2}j_{k_2+1}})
    (\sum_{(t_0 \cdots t_{l-1}) \in \atop \hat \sZ(l-1)} h_{j_l, t_0}^0\prod_{k_3=0}^{l-1}w_{t_{k_3}, t_{k_3+1}}^k)
    \label{eq:wpaths}
\end{equation}

Let's denote $\prod_{k_1=0}^{l-1}\delta(\hat h_{j_{k_1},t}^{l-k_1-1})$ as $\delta_p$, and $\prod_{k_2=0}^{l-1}d_{j_{k_2}j_{k_2+1}}$ as $d_p$, then we have:

\begin{equation}
    var([h_i^l]_{p,t}) = d_p^2\ var(\delta_p \sum_{(t_0 \cdots t_{l-1}) \in \atop \hat \sZ(l-1)} h_{j_l, t_0}^0\prod_{k=0}^{l-1}w_{t_{k}, t_{k+1}}^k)
    \label{eq:neu_p}
\end{equation}

Let's recap the expression of weight propagation path:

\begin{equation}
[h_i^l]_{p,t,\phi}=h_{j_l, t_0}^0 w_{t_0, t_1}^0 \cdots w_{t_{l-1}, t}^{l-1} 
\end{equation}

It's obvious that $\sum_{(t_0 \cdots t_{l-1}) \in \atop \hat \sZ(l-1)} h_{j_l, t_0}^0\prod_{k=0}^{l-1}w_{t_{k}, t_{k+1}}^k$ in~\cref{eq:neu_p} is the sum over $[h_i^l]_{p,t,\phi}$. We now derive the variance of $\delta_p[h_i^l]_{p,t,\phi}$:

\begin{equation}
    var(\delta_p[h_i^l]_{p,t,\phi}) = (h_{j_l, t_0}^0)^2\ var(\delta_p \prod_{k=0}^{l-1} w^k_{t_k, t_{k+1}})
    \label{eq:path}
\end{equation}

where

\begin{subequations}
\begin{align}
    var(\delta_p \cdot \prod_{k=0}^{l-1} w^k_{t_k, t_{k+1}}) &= cov[\delta_p^2, \prod_{k=0}^{l-1} (w^k_{t_k, t_{k+1}})^2]\\
    &+[var(\delta_p)+E^2(\delta_p)]\cdot[var(\prod_{k=0}^{l-1} w^k_{t_k, t_{k+1}})+E^2(\prod_{k=0}^{l-1} w^k_{t_k, t_{k+1}})]\\
    &-[cov(\delta_p, \prod_{k=0}^{l-1} w^k_{t_k, t_{k+1}})+E(\delta_p)E(\prod_{k=0}^{l-1} w^k_{t_k, t_{k+1}})]^2
\end{align}
\end{subequations}

$w^k_{t_k, t_{k+1}}$ for different $k$ are independent and have the same mean value 0, thus both $E(\prod_{k=0}^{l-1} w^k_{t_k, t_{k+1}}) = \prod_{k=0}^l E(w^k_{t_k, t_{k+1}})$ and $E^2(\prod_{k=0}^{l-1} w^k_{t_k, t_{k+1}})$ are equal to 0, $var(\prod_{k=0}^{l-1} w^k_{t_k, t_{k+1}}) = \prod_{k=0}^l (var(w^k_{t_k, t_{k+1}})+E^2(w^k_{t_k, t_{k+1}})) - \prod_{k=0}^l(E^2(w^k_{t_k, t_{k+1}}))$ is equal to $\prod_{k=0}^{l-1} var(w^k_{t_k, t_{k+1}})$. Additionally, given that~\cref{assu:delta} holds, $\delta_p$ is a Bernoulli random variable and has the success probability $0.5^l$, where $l$ is the length of $p$. Furthermore, when~\cref{assu:deltawpath} holds, the correlation between $\delta_p$ and $\prod_{k=0}^{l-1} w^k_{t_k, t_{k+1}}$, between $\delta_p^2$ and $\prod_{k=0}^{l-1} (w^k_{t_k, t_{k+1}})^2$, are both approximately equal to 0. As a result, we have:

\begin{equation}
    var(\delta_p\prod_{k=0}^{l-1} w^k_{t_k, t_{k+1}}) = 0.5^l\cdot\prod_{k=0}^{l-1} var(w^k_{t_k, t_{k+1}})
    \label{eq:no_name_1}
\end{equation}

Combining~\cref{eq:no_name_1,eq:path}, we have:

\begin{equation}
    var(\delta_p[h_i^l]_{p,t,\phi})=(h_{j_l, t_0}^0)^20.5^l\cdot\prod_{k=0}^{l-1} var(w^k_{t_k, t_{k+1}})
    \label{eq:var_path}
\end{equation}

When~\cref{assu:wpaths} holds,~\cref{eq:neu_p} is equal to:

\begin{subequations}
\begin{align}
    var([h_i^l]_{p,t}) &= d_p^2 \sum_{\phi \in \Phi_{i,p,t}} var(\delta_p[h_i^l]_{p,t,\phi}) \\
    &= d_p^2\ 0.5^l\cdot\prod_{k=0}^{l-1} var(w^k_{t_k, t_{k+1}}) \cdot \sum_{\phi \in \Phi_{i,p,t}} (h_{j_l, t_0}^0)^2
\end{align}
\end{subequations}

$var(h_{i,t}^l)$ is equal to the summation of $var([h_{i,t}^l]_{p,t})$ for all $p$ in $\sP_{i,l}$, thus we have:

\begin{subequations}
\begin{align}
    var(h_{i,t}^l) &=  var(\sum_{p \in \sP_{i, l}}var([h_i^l]_{p,t}))\\
    &= \sum_{p \in \sP_{i, l}}var([h_i^l]_{p,t}) + 2\sum_{i < j}cov([h_i^l]_{p_i,t}, [h_i^l]_{p_j,t})
\end{align}
\end{subequations}

where

\begin{equation}
cov([h_i^l]_{p_i,t}, [h_i^l]_{p_j,t}) = d_{p_i} d_{p_j} cov(\sum_{\phi \in \Phi_{i, p, t}}\delta_p[h_i^l]_{p,t,\phi}, \sum_{\phi \in \Phi_{j, p, t}}\delta_p[h_j^l]_{p,t,\phi})
\end{equation}

where

\begin{subequations}
\begin{align}
&cov(\sum_{\phi \in \Phi_{i, p, t}}\delta_p[h_i^l]_{p,t,\phi}, \sum_{\phi \in \Phi_{j, p, t}}\delta_p[h_j^l]_{p,t,\phi}) \\
&= cov(\sum_{\phi \in \Phi_{i, p, t}}\delta_p h_{j_l, t_0}^0\prod_{k=0}^{l-1} var(w^k_{t_k, t_{k+1}}), \sum_{\phi \in \Phi_{j, p, t}}\delta_{p_j} h_{j_l', t_0}^0\prod_{k=0}^{l-1} var(w^k_{t_k, t_{k+1}}))\\
&= h_{j_l, t_0}^0 h_{j_l', t_0}^0 cov(\sum_{\phi \in \Phi_{i, p, t}}\delta_p \prod_{k=0}^{l-1} var(w^k_{t_k, t_{k+1}}), \sum_{\phi \in \Phi_{j, p, t}}\delta_{p_j} \prod_{k=0}^{l-1} var(w^k_{t_k, t_{k+1}}))
\end{align}
\end{subequations}

When~\cref{assu:1} holds, the Pearson correlation between $\sum_{\phi} \delta_p \prod_{k=0}^{l-1} var(w^k_{t_k, t_{k+1}})$ and $\sum_{\phi} \delta_{p_j} \prod_{k=0}^{l-1} var(w^k_{t_k, t_{k+1}})$ is close to 1, we thus use $\sqrt{var(\sum_{\phi} \delta_p \prod_{k=0}^{l-1} var(w^k_{t_k, t_{k+1}}))var(\sum_{\phi} \delta_{p_j} \prod_{k=0}^{l-1} var(w^k_{t_k, t_{k+1}}))}$ to approximate the covariance term. And according to the analysis above, the correlation between different weight propagation paths is approximately 0, thus the covariance term can be approximated by $\sqrt{\sum_{\phi} var(\delta_p \prod_{k=0}^{l-1} var(w^k_{t_k, t_{k+1}}))\sum_{\phi} var( \delta_{p_j} \prod_{k=0}^{l-1} var(w^k_{t_k, t_{k+1}}))} = \sum_{\phi} var(\delta_p \prod_{k=0}^{l-1} var(w^k_{t_k, t_{k+1}}))$. There are $\prod_{l'=0}^{l-1} m_1^{(l')}$ weight propagation paths in $\Phi_{i,p,t}$ for $L$ layers. We thus replace $\sum_{\phi} var(\delta_p \prod_{k=0}^{l-1} var(w^k_{t_k, t_{k+1}}))$ with $\prod_{l'=0}^{l-1} m_1^{(l')} \cdot var(\delta^l \prod_{k=0}^{l-1} var(w^k_{t_k, t_{k+1}}))$. As a result, with $\hat h_{j_l, t_0}^0$ to denote the mean over elements of $\vh_{np_i}^0$we have:

\begin{align}
    var(h_{i,t}^l) &= \sum_{p \in \sP_{i, l}}(\prod_{j, j' \in p}d_{jj'})^2 0.5^l\cdot\prod_{k=0}^{l-1} var(w^k_{t_k, t_{k+1}}) \sum_{\phi \in \Phi_{i,p,t}} (\hat h_{j_l, t_0}^0)^2 \\
    &+ 2m^l \cdot 0.5^l\cdot\prod_{k=0}^{l-1} var(w^k_{t_k, t_{k+1}}) \sum_{i<j \atop p_i, p_j \in \sP_{i, l}}(\prod_{t, t' \in p_i}d_{tt'} \prod_{t, t' \in p_j}d_{tt'})(\hat h_{j_l, t_0}^0 \hat h_{j_l', t_0}^0)
    \label{eq:varhi_1}
\end{align}

where

\begin{align}
    \sum_{p \in \sP_{i, l}}(\prod_{j, j' \in p}d_{jj'})^2\sum_{\phi \in \Phi_{i,p,t}} (\hat h_{j_l, t_0}^0)^2 = \prod_{l'=0}^{l-1} m_1^{(l')} \sum_{p \in \sP_{i, l}}(\prod_{j, j' \in p}d_{jj'}\hat h_{j_l, t_0}^0)^2 \label{eq:temp_1}
\end{align}

and 
\begin{align}
    &2\sum_{i<j \atop p_i, p_j \in \sP_{i, l}}(\prod_{t, t' \in p_i}d_{tt'} \prod_{t, t' \in p_j}d_{tt'})(\hat h_{j_l, t_0}^0 \hat h_{j_l', t_0}^0)\\
    =&\sum_{i \ne j \atop p_i, p_j \in \sP_{i, l}}(\prod_{t, t' \in p_i }d_{tt'}\hat h_{j_l, t_0}^0)(\prod_{t, t' \in p_j}d_{tt'}\hat h_{j_l', t_0}^0) \\
    =&\sum_{p_i \in \sP_{i, l}}(\prod_{t, t' \in p_i}d_{tt'}\hat h_{j_l, t_0}^0)\sum_{p_j \in \sP_{i, l}}(\prod_{t, t' \in p_j}d_{tt'}\hat h_{j_l', t_0}^0) - \sum_{p_i' \in \sP_{i, l}}(\prod_{t, t' \in p_i'}d_{tt'}\hat h_{j_l'',t_0}^0)^2 \label{eq:temp_2}
\end{align}

Note that in \Eqref{eq:temp_2}, $\sum_{p_i \in \sP_{i, l}}(\prod_{t, t' \in p_i}d_{tt'}\hat h_{j_l, t_0}^0)$ is equal to $\sum_{p_j \in \sP_{i, l}}(\prod_{t, t' \in p_j}d_{tt'}\hat h_{j_l', t_0}^0)$, and $\sum_{p_i' \in \sP_{i, l}}(\prod_{t, t' \in p_i'}d_{tt'}\hat h_{j_l'', t_0}^0)^2$ of \Eqref{eq:temp_2} is equal to $\sum_{p \in \sP_{i, l}}(\prod_{j, j' \in p}d_{jj'}\hat h_{j_l, t_0}^0)^2$ of \Eqref{eq:temp_1}, thus for \Eqref{eq:varhi_1} we have: 

\begin{align}
    var(h_{i,t}^l) &= \prod_{l'=0}^{l-1} m_1^{(l')} \cdot 0.5^l \cdot \prod_{k=0}^{l-1} var(w^k_{t_k, t_{k+1}}) \cdot \sum_{p \in \sP_{i, l}}(\prod_{j, j' \in p}d_{jj'}\hat h_{j_l, t_0}^0)^2 \\
    &+\prod_{l'=0}^{l-1} m_1^{(l')} \cdot 0.5^l \cdot \prod_{k=0}^{l-1} var(w^k_{t_k, t_{k+1}}) \cdot \sum_{p_i \in \sP_{i, l}}(\prod_{t, t' \in p_i}d_{tt'}\hat h_{j_l, t_0}^0)\sum_{p_j \in \sP_{i, l}}(\prod_{t, t' \in p_j}d_{tt'}\hat h_{j_l', t_0}^0)\\
    &-\prod_{l'=0}^{l-1} m_1^{(l')} \cdot 0.5^l \cdot \prod_{k=0}^{l-1} var(w^k_{t_k, t_{k+1}}) \cdot \sum_{p_i' \in \sP_{i, l}}(\prod_{t, t' \in p_i'}d_{tt'}\hat h_{j_l'',t_0}^0)^2\\
    &=\prod_{l'=0}^{l-1} m_1^{(l')} \cdot 0.5^l \cdot \prod_{k=0}^{l-1} var(w^k_{t_k, t_{k+1}}) \cdot [\sum_{p \in \sP_{i, l}}(\prod_{t, t' \in p}d_{tt'}\hat h_{j_l'',t_0}^0)]^2\\
    &=\prod_{l'=0}^{l-1} m_1^{(l')} \cdot 0.5^l \cdot \prod_{k=0}^{l-1} var(w^k_{t_k, t_{k+1}}) \cdot [\widetilde \mA^l\vh^0]^2_i
\end{align}

where $[\vv]^2_i$ denotes the $i_{th}$ element of the element-wise square of the vector $\vv$, $\widetilde \mA$ is the renormalized adjacent matrix of the GCN, and $\vh^0$ is a vector equal to $[h_0^0, h_1^0, \cdots, h_{|\sN|-1}^0]^T$, where $h_i^0$ is the mean over elements of $\vh_i^0$. Note that $\widetilde \mA^l\vh^0$ is the message passing of the GCN with the input as a input graph and mean of node features. In practice, we use GCN implemented by DGL to accelerate computing $\widetilde \mA^l\vh^0$. Let $\vi \in \sR^{|\sN|\times 1}$ be a vector with all ones, we intend to make $\sum_i var(h_{i,t}^l)$ equal to $\sum_i var(h_{i,t}^{l+1})$. Thus we have:

\begin{align}
    \sum_i var(h_{i,t}^l) &= \sum_i var(h_{i,t}^{l+1})\\
    \prod_{l'=0}^{l-1} m_1^{(l')} \cdot 0.5^l \cdot \prod_{k=0}^{l-1} var(w^k_{t_k, t_{k+1}}) \cdot \vi^T[\widetilde \mA^l\vh^0]^2 &= \prod_{l'=0}^{l} \cdot 0.5^{l+1} \cdot \prod_{k=0}^l var(w^k_{t_k, t_{k+1}}) \cdot \vi^T[\widetilde \mA^{l+1}\vh^0]^2\\
    var(w^l) &= \frac{2}{m_1^{(l)}} \frac{\vi^T[\widetilde \mA^{l-1}\vh^0]^2}{\vi^T [\widetilde \mA^l\vh^0]^2}
    \label{eq:w_for}
\end{align}

\textbf{\cref{assu:1c}}\ \ We use the same experiment setting as in the empirical verification of~\cref{assu:1}, and calculate the mean and standard deviation of neurons of the softmax of $h_i^L$. In~\cref{tab:est_1c}, we present evaluated results in the row titled with \textbf{Evaluated}. The row titled with \textbf{Expected} indicates that we require the mean of evaluated results to be expected numbers to hold~\cref{assu:1c}. We observe that evaluated results are quite close to expected numbers.

\begin{table}[htbp]
\vspace{-0.3cm}
 \centering
 \caption{Estimation of the success probability of $\delta_p$.} 
 \label{tab:est_1c}
 \vskip 0.1in
\begin{tabular}{ccccccc}
\toprule 
& \multicolumn{4}{c}{Datasets} & \\ & cora & pubmed & citeseer & arxiv\\
\midrule
\midrule
Evaluated & 0.14±0.01 & 0.34±0.02 & 0.16±0.01 & 0.024±0.02\\
\midrule
Expected & 0.14 & 0.33 & 0.17 & 0.025\\
\bottomrule
\end{tabular}
\end{table}

\paragraph{Proof of Theorem 2} Let's define the loss function as $Loss$, and $L$ as the number of layers. The backward gradients of $Loss$ w.r.t $\vh_i^l$ is $\frac{\partial Loss}{\partial \vh_i^l}$, of which elements are assumed to be i.i.d. Thus we only consider the variance of $\frac{\partial Loss}{\partial h_{i,k}^l}$, where $h_{i,k}^l$ can be any element of $\vh_i^l$. The variance of $\frac{\partial Loss}{\partial h_{i,k}^l}$ is:

\begin{align}
    \frac{\partial Loss}{\partial h_{i,k}^l} = \sum_{t \in \sN, k'} \frac{\partial Loss}{\partial h_{t, k'}^L} \frac{\partial h_{t, k'}^L}{\partial h_{i, k}^l}
\end{align}

where

\begin{align}
    \frac{\partial h_{t, k'}^L}{\partial h_{i, k}^l} = \sum_{p \in \sP_{i^l, t^L}} \frac{\partial [h_{t, k'}^L]_p}{\partial h_{i, k}^l}
\end{align}

where $\sP_{i^l, t^L}$ denote the set of all message propagation paths from $\vh_i^l$ to $\vh_t^L$. We firstly look at $\frac{\partial h_{t, k'}^L}{\partial h_{i, k}^l}$. $[h_{t, k'}^L]_p$ is equal to $\delta_p^{L-l} \prod_{j, j' \in p}d_{jj'}  \sum_{k_{L-l}}\sum_{k_{L-2}}\cdots\sum_{k_{l+1}}h_{i, k}^lw_{k, k_{l+1}}^lw_{k_{l+1}, k_{l+2}}^{l+1}\cdots w_{k_{L-1}, k'}^{L-1}$, Thus we have:

\begin{align}
    \frac{\partial [h_{t, k'}^L]_p}{\partial h_{i, k}^l} = \delta_p^{L-l} \prod_{j, j' \in p}d_{jj'}  \sum_{k_{L-l}}\sum_{k_{L-2}}\cdots\sum_{k_{l+1}}w_{k, k_{l+1}}^lw_{k_{l+1}, k_{l+2}}^{l+1}\cdots w_{k_{L-1}, k'}^{L-1}
\end{align}

and 

\begin{align}
    var(\frac{\partial [h_{t, k'}^L]_p}{\partial h_{i, k}^l}) = m^{L-l-1}(\prod_{j, j' \in p}d_{jj'})^2\rho_{L-l}\cdot\prod_{k=l+1}^{L-1} var(w^k_{t_k, t_{k+1}})
\end{align}

where the $m$ above is the \textit{fan\_out} of each layer, which is different from the one defined in in proof of Theorem 1. Similar to \Eqref{eq:varhi_1} and its subsequent derivations, we have:

\begin{align}
    var(\frac{\partial h_{t, k'}^L}{\partial h_{i, k}^l}) = m^{L-l-1} \cdot \rho_{L-l} \cdot \prod_{k=l+1}^{L-1} var(w^k_{t_k, t_{k+1}}) \cdot \vi^T[\widetilde \mA^{L-l}\vi]^2
\end{align}

Assuming that the loss function is cross entropy, and the label of node $t$ is $y(t)$, we have:

\begin{equation}
\frac{\partial Loss}{\partial h_{t, k'}^L}=
\begin{cases}
\frac{s(\vh_t^L)_{k'}-1}{|\sN|}& k'=y(t)\\
\frac{s(\vh_t^L)_{k'}}{|\sN|}& k' \ne y(t)
\end{cases}
\label{eq:grad_last}
\end{equation}

where $s(\vv)_i$ is the $i_{th}$ element of the \textit{softmax} of a vector $\vv$. \textbf{Experiments} show that at the first epoch, $s(\vh_t^L)_{k'}$ is approximately $1/C$, where $C$ is the number of classes. Thus we have:

\begin{align}
    var(\frac{\partial Loss}{\partial h_{i,k}^l}) &= var(\sum_{t \in \sN, k'} \frac{\partial Loss}{\partial h_{t, k'}^L} \frac{\partial h_{t, k'}^L}{\partial h_{i, k}^l})\\
    &=\sum_{t \in \sN, k'} (\frac{\partial Loss}{\partial h_{t, k'}^L})^2 var(\frac{\partial h_{t, k'}^L}{\partial h_{i, k}^l})\\
    &=\frac{1}{|\sN|^2}(1-\frac{1}{C}) \cdot m^{L-l-1} \cdot \rho_{L-l} \cdot \prod_{k=l+1}^{L-1} var(w^k_{t_k, t_{k+1}}) \cdot [\widetilde \mA^{L-l}\vi]^2_i
\end{align}

Similar to the proof of Theorem 1, we intend to make $\sum_i var(\frac{\partial Loss}{\partial h_{i,k}^l})$ equal to $\sum_i var(\frac{\partial Loss}{\partial h_{i,k}^{l+1}})$, thus we have:

\begin{align}
    &\frac{1}{|\sN|^2}(1-\frac{1}{C}) \cdot m^{L-l-1} \cdot \rho_{L-l} \cdot \prod_{k=l+1}^{L-1} var(w^k_{t_k, t_{k+1}}) \cdot \vi^T[\widetilde \mA^{L-l}\vi]^2 \\
    =&\frac{1}{|\sN|^2}(1-\frac{1}{C}) \cdot m^{L-l-2} \cdot \rho_{L-l-1} \cdot \prod_{k=l+2}^{L-1} var(w^k_{t_k, t_{k+1}}) \cdot \vi^T[\widetilde \mA^{L-l-1}\vi]^2
\end{align}

Then we have:

\begin{equation}
    var(w^k_{t_k, t_{k+1}})=\frac{2}{m_2^{k}}\frac{\vi^T[\widetilde \mA^{L-l-1}\vi]^2}{\vi^T[\widetilde \mA^{L-l}\vi]^2}
\end{equation}

For the last layer $var(w^{L-1})$, we obtain it by making $\sum_i var(\frac{\partial Loss}{\partial h_{i,k}^L})$ equal to $\sum_i var(\frac{\partial Loss}{\partial h_{i,k}^{L-1}})$. According to \Eqref{eq:grad_last}, the $var(\frac{\partial Loss}{\partial h_{i,k}^L})$ is approximately equal to $1/(|\sN|^2C)$. Thus we have:

\begin{align}
    \frac{|\sN|}{|\sN|^2C}
    =&\frac{1}{|\sN|^2}(1-\frac{1}{C}) \cdot var(w^{L-1}) \cdot \vi^T[\widetilde \mA\vi]^2\\
    var(w^{L-1})=&\frac{|\sN|}{(C-1)\vi^T[\widetilde \mA\vi]^2}
\end{align}

\section{Experimental Setting}
\label{app_exp}

In this section, we present design space of hyperparameter tunning used in~\cref{sec:exp}. 

\subsection{Node Classification}
For node classification tasks, we perform a grid search to tune the hyperparameters. We list the hyperparameter tunning settings in~\cref{tab:nchyper}, where lr and wd denotes the learning rate and the weight decay, respectively, and num\_layers represents the number of GNN layers. The patience are used for early-stopping.

\begin{table*}[htbp]
    \centering
    \caption{The hyperparameter tunning setting of experiments in~\cref{tab:nc}.}
    \label{tab:nchyper}
    \begin{tabular}{ cl }
    \toprule 
        Datasets & Hyper-parameters \\
    \midrule
        cora & 
        \makecell[l]{
            hidden\_channels: \{32, 64, 128\}, 
            num\_layers: \{2, 3, 4\}, 
            lr: \{1e-3, 5e-3, 1e-2, 5e-2\}, \\
            epochs: 1000, 
            patience: 20, 
            dropout: \{0.0, 0.5\}, 
            wd: \{0.0, 5e-6\}
        } \\
    \midrule
        pubmed &
        \makecell[l]{
            hidden\_channels: \{32, 64, 128\}, 
            num\_layers: \{2, 3, 4\}, 
            lr: \{1e-3, 5e-3, 1e-2, 5e-2\}, \\
            epochs: 1000, 
            patience: 20, 
            dropout: \{0.0, 0.5\}, 
            wd: \{0.0, 5e-6\}
        } \\
    \midrule
        citeseer &
        \makecell[l]{
            hidden\_channels: \{32, 64, 128\}, 
            num\_layers: \{2, 3, 4\}, 
            lr: \{1e-3, 5e-3, 1e-2, 5e-2\}, \\
            epochs: 1000, 
            patience: 20, 
            dropout: \{0.0, 0.5\}, 
            wd: \{0.0, 5e-6\}
        } \\
    \midrule
        arxiv &
        \makecell[l]{
            hidden\_channels: \{128, 256\}, 
            num\_layers: \{3, 4, 5\}, 
            lr: \{1e-3, 5e-3, 1e-2, 5e-2\}, \\
            epochs: 1000, 
            patience: 100, 
            wd: \{0.0, 5e-6, 5e-5\}, 
            dropout: \{0.0, 0.5\}
        } \\
    \midrule
        proteins &
        \makecell[l]{
            hidden\_channels: \{128, 256\}, 
            num\_layers: 3, 
            lr: \{1e-3, 5e-3, 1e-2, 5e-2\}, \\
            dropout: \{0.0, 0.5\}, 
            wd: \{0.0, 5e-6\}, 
            epochs: 2000
        } \\
    \midrule
        products &
        \makecell[l]{
            hidden\_channels: 256, 
            num\_layers: \{3, 4\}, 
            lr: \{1e-3, 5e-3\}, \\
            dropout: 0.5, 
            wd: \{0.0, 5e-5\}, 
            epochs: 100
        } \\
    \bottomrule
    \end{tabular}
\end{table*}

\subsection{Link Prediction}
For all OGB datasets, we follow the rules of OGB and employ the same data splitting. We use the Adam optimizer with zero weight\_decay. We list the hyperparameter tunning settings for the link prediction tasks in~\cref{tab:lphyper}, where lr represents the learning rate and num\_layers represents the number of GNN layers. For NGNN models, we use ngnn\_type to denote the position where the non-linear layers are inserted, (for instance, \textit{input} means applying NGNN to only the input GNN layer), and use num\_ngnn\_layers to denote the number of nonlinear layers in each NGNN block. Since the citation2 datasets is too large, we use the memory-friendly variants of (NGNN-)GCN and (NGNN-)GraphSAGE, namely ClusterGCN and Neighbor Sampling(SAGE aggregation).

\begin{small}
\begin{table}[htbp]
    \centering
    \caption{The hyperparameter tunning setting of experiments in~\cref{tab:lp}.}
    \label{tab:lphyper}
    \begin{tabular}{ cll }
    \toprule 
        Datasets & Models & Hyper-parameters \\
    \midrule
        \multirow{4}{*}[-11.44ex]{ddi}
            & GCN & 
            \makecell[l]{
                lr: \{0.002, 0.005\}, 
                batch\_size: \{16384, 32768\}, \\
                dropout: \{0.4, 0.5, 0.6\}, 
                epoch: \{1600, 2400\}, \\
                num\_layers: 2, 
                hidden\_channels: 256 
            } \\ \Xcline{2-3}{0.5pt} \specialrule{0em}{1pt}{1pt}
            & NGNN-GCN & 
            \makecell[l]{
                ngnn\_type: \{input, all\}, 
                num\_ngnn\_layers: 1, \\
                lr: \{0.001, 0.0015, 0.002, 0.005\}, 
                batch\_size: \{8192, 16384, 32768\}, \\
                dropout: \{0.2, 0.3, 0.4, 0.5, 0.6, 0.7\}, 
                epoch: \{800, 1600, 2400\}, \\
                num\_layers: 2, 
                hidden\_channels: 256 
            } \\ \Xcline{2-3}{0.5pt} \specialrule{0em}{1pt}{1pt}
            & GraphSAGE & 
            \makecell[l]{
                lr: \{0.001, 0.002\}, 
                batch\_size: 32768, \\
                dropout: \{0.1, 0.2, 0.3, 0.4\}, 
                epoch: \{800, 1200, 2000\}, \\
                num\_layers: 2, 
                hidden\_channels: 256
            } \\ \Xcline{2-3}{0.5pt} \specialrule{0em}{1pt}{1pt}
            & NGNN-GraphSAGE & 
            \makecell[l]{
                ngnn\_type: input, 
                num\_ngnn\_layers: 1, \\
                lr: \{0.0005, 0.001, 0.005\}, 
                batch\_size: \{8192, 16384, 32768\}, \\
                dropout: \{0, 0.1, 0.2, 0.3, 0.4\}, 
                epoch: \{600, 1200, 2000\}, \\
                num\_layers: 2, 
                hidden\_channels: 256
            } \\
    \midrule
        \multirow{4}{*}[-10.16ex]{collab} 
            & GCN & 
            \makecell[l]{
                lr: \{0.0005, 0.001\}, 
                batch\_size: \{32768, 65536\}, 
                dropout: \{0.2, 0.3\}, \\
                epoch: \{800, 1200\}, 
                num\_layers: \{3, 4\}, 
                hidden\_channels:256
            } \\ \Xcline{2-3}{0.5pt} \specialrule{0em}{1pt}{1pt}
            & NGNN-GCN & 
            \makecell[l]{
                ngnn\_type: \{input, hidden, all\}, 
                num\_ngnn\_layers: 2, \\
                lr: \{0.0005, 0.001, 0.002, 0.005\}, 
                batch\_size: \{32768, 65536\}, \\
                dropout: \{0.2, 0.3\}, 
                epoch: \{800, 1200\}, \\
                num\_layers: \{3, 4\}, 
                hidden\_channels:256
            } \\ \Xcline{2-3}{0.5pt} \specialrule{0em}{1pt}{1pt}
            & GraphSAGE & 
            \makecell[l]{
                lr: \{0.0005, 0.001\}, 
                batch\_size: \{65536, 131072\}, 
                dropout: 0.2, \\
                epoch: \{600, 900\}, 
                num\_layers: 4, 
                hidden\_channels: 256
            } \\ \Xcline{2-3}{0.5pt} \specialrule{0em}{1pt}{1pt}
            & NGNN-GraphSAGE & 
            \makecell[l]{
                ngnn\_type: \{input, hidden, all\}, 
                num\_ngnn\_layers: 2, \\
                lr: \{0.0005, 0.001, 0.002\}, 
                batch\_size: \{32768, 65536, 131072\}, \\
                dropout: \{0, 0.1, 0.2, 0.3\}, 
                epoch: \{400, 800, 1200\}, \\
                num\_layers: \{3, 4\}, 
                hidden\_channels: 256
            } \\
    \midrule
        \multirow{4}{*}[-9.79ex]{citation2} 
            & GCN & 
            \makecell[l]{
                lr: \{0.0003, 0.0005, 0.001\}, 
                batch\_size: 256, 
                dropout: \{0, 0.1, 0.2\}, \\
                epoch: 200, 
                num\_layers: 3, 
                hidden\_channels: 256
            } \\ \Xcline{2-3}{0.5pt} \specialrule{0em}{1pt}{1pt}
            & NGNN-GCN & 
            \makecell[l]{
                ngnn\_type: \{hidden, input\}, 
                num\_ngnn\_layers: \{1, 2\}, \\
                lr: \{0.0003, 0.0005\}, 
                batch\_size: 256, \\
                dropout: \{0, 0.1\}, 
                epoch: 200, 
                eval\_step: 10, \\
                num\_layers: 3, 
                hidden\_channels: 256
            } \\ \Xcline{2-3}{0.5pt} \specialrule{0em}{1pt}{1pt}
            & GraphSAGE & 
            \makecell[l]{
                lr: \{0.0003, 0.0005, 0.001\}, 
                batch\_size: 1024, 
                dropout: \{0, 0.2\}, \\
                epoch: 200, 
                num\_layers: 3, 
                hidden\_channels:256
            } \\ \Xcline{2-3}{0.5pt} \specialrule{0em}{1pt}{1pt}
            & NGNN-GraphSAGE & 
            \makecell[l]{
                ngnn\_type: \{hidden, input\}, 
                num\_ngnn\_layers: \{1, 2\}, \\
                lr: \{0.0003, 0.0005, 0.001\}, 
                batch\_size: 1024, 
                dropout: \{0, 0.2\}, \\
                epoch: 200, 
                num\_layers: 3, 
                hidden\_channels: 256
            } \\
    \midrule
        \multirow{4}{*}[-7.12ex]{ppa} 
            & GCN & 
            \makecell[l]{
                lr: 0.001, 
                batch\_size: \{32768, 49152, 65536\}, \
                dropout: \{0.2, 0.3\}, \\
                epoch: \{120, 150\}, 
                num\_layers: \{3, 4\}, 
                hidden\_channels: 256
            } \\ \Xcline{2-3}{0.5pt} \specialrule{0em}{1pt}{1pt}
            & NGNN-GCN & 
            \makecell[l]{
                ngnn\_type: input, 
                num\_ngnn\_layers: 2, 
                lr: 0.001, \\
                batch\_size: \{32768, 49152, 65536\}, 
                dropout: \{0.2, 0.3\}, \\
                epoch: \{120, 150\}, 
                num\_layers: \{3, 4\}, 
                hidden\_channels: 256
            } \\ \Xcline{2-3}{0.5pt} \specialrule{0em}{1pt}{1pt}
            & GraphSAGE & 
            \makecell[l]{
                lr: \{0.001, 0.0015\}, 
                batch\_size: \{49152, 65536\}, 
                dropout: 0.2, \\
                epoch: \{120, 150\}, 
                num\_layers: 4, 
                hidden\_channels: 256
            } \\ \Xcline{2-3}{0.5pt} \specialrule{0em}{1pt}{1pt}
            & NGNN-GraphSAGE & 
            \makecell[l]{
                ngnn\_type: input, 
                num\_ngnn\_layers: 2, 
                lr: \{0.001, 0.0015\}, \\
                batch\_size: \{49152, 65536\}, 
                dropout: 0.2, 
                epoch: \{120, 150\}, \\
                num\_layers: \{3, 4\}, 
                hidden\_channels: 256
            } \\
    \bottomrule
    \end{tabular}
\end{table}
\end{small}

\subsection{Graph Classification}
For graph classification tasks, the grid search settings for hyperparameter tunning are listed in~\cref{tab:gchyper}, where lr and wd denotes the learning rate and the weight decay, respectively, and num\_layers represents the number of GNN layers. For NGNN variants, ngnn\_type refers to the position of the linear layer. For example, \textit{input} means there is one linear layer stacked on the first GNN layer, and \textit{all} means each GNN layer is equipped with a linear layer. num\_ngnn\_layers refers to the number of linear layers in the GNN model. Specific details can be found in~\cite{ngnn}.

\begin{table}[htbp]
    \centering
    \caption{The hyperparameter tunning setting of experiments in~\cref{tab:gc}.}
    \label{tab:gchyper}
    \begin{tabular}{ cl }
    \toprule 
        Datasets & Hyper-parameters \\
    \midrule
        molhiv & 
        \makecell[l]{
		lr: \{5e-4, 1e-4, 5e-3, 1e-3\}, 
            dropout: \{0.0, 0.5\}, 
            wd: \{0.0, 5e-6\}, \\
		batch\_size: 32, 
		hidden\_channels: 300, 
		num\_layer: 5
        } \\
    \midrule
        molpcba &
        \makecell[l]{
		lr: \{5e-4, 1e-4, 5e-3, 1e-3\}, 
            dropout: \{0.0, 0.5\}, 
            wd: \{0.0, 5e-6\}, \\
		batch\_size: 32, 
		hidden\_channels: 300, 
		num\_layer: 5
        } \\
    \bottomrule
    \end{tabular}
\end{table}

\section{Miscellaneous}
\label{misc}
We present intermediate steps to derive~\cref{eq:mpath_ori} in following equations.
\begin{equation}
\begin{aligned}
    \vh_i^{l}=&\delta(\hat \vh_i^{l-1}) \odot \hat \vh_i^{l-1}\\
    =&\delta(\hat \vh_i^{l-1}) \odot \sum_{j_1 \in \sN(i)}d_{ij_1}\delta(\hat \vh_{j_1}^{l-1}) \odot \hat \vh_{j_1}^{l-1} \cdot \mW^{l-1}\\
    =&\delta(\hat \vh_i^{l-1}) \odot \sum_{j_1 \in \sN(i)}d_{ij_1}\delta(\hat \vh_{j_1}^{l-1}) \odot
    \sum_{j_2 \in \sN(j_1)}d_{j_1j_2}\delta(\hat \vh_{j_2}^{l-2})\\
    &\odot \hat \vh_{j_2}^{l-2} \cdot \mW^{l-2}\mW^{l-1}\\
    &\cdots\\
    =&\delta(\hat \vh_i^{l-1}) \odot (\prod_{k_1=0 \atop \odot}^{l-2}\sum_{j_{k_1+1} \in \atop \sN(j_{k_1})}d_{j_{k_1}j_{k_1+1}}
    \delta(\hat \vh_{j_{k_1+1}}^{l-k_1-2}))\\
    &\odot(\sum_{j_l \in \sN(j_{l-1})}d_{j_{l-1}j_l}\vh_{j_l}^0)\cdot(\prod_{k_2=0}^{l-1}\mW^{k_2})
\end{aligned}
\end{equation}

\end{document}